\documentclass{article}

\usepackage{arxiv}

\usepackage[utf8]{inputenc} 
\usepackage[T1]{fontenc}    
\usepackage{url}            
\usepackage{booktabs}       
\usepackage{amsfonts}       
\usepackage{nicefrac}       
\usepackage{microtype}      
\usepackage{lipsum}		
\usepackage{graphicx}
\usepackage{natbib}
\usepackage{doi}
\usepackage{amsmath}
\usepackage{algorithm}
\usepackage[noend]{algpseudocode}
\usepackage{multirow}
\hypersetup{hidelinks}
\usepackage{wrapfig}
\usepackage{svg}
\usepackage{tabularx}
\usepackage{bm}
\usepackage{multicol}
\usepackage{hyperref}
\usepackage{caption}

\usepackage{caption}

\title{Leveraging Constraint Programming in a Deep Learning Approach for Dynamically Solving the Flexible Job-Shop Scheduling Problem}

\author{Imanol Echeverria\thanks{Corresponding author.} \\
    TECNALIA, Basque Research and Technology Alliance (BRTA) \\
	\texttt{imanol.echeverria@tecnalia.com} \\
	\And
	Maialen Murua\\
	TECNALIA, Basque Research and Technology Alliance (BRTA)\\
	\texttt{maialen.murua@tecnalia.com} \\
 	\And
	Roberto Santana\\
	Computer Science and Artificial Intelligence Department, University of the Basque Country\\
	\texttt{roberto.santana@ehu.eus} \\
}

\date{}

\renewcommand{\headeright}{}
\renewcommand{\undertitle}{}
\renewcommand{\shorttitle}{}

\hypersetup{
pdftitle={Leveraging Constraint Programming in a Deep Learning Approach for Dynamically Solving the Flexible Job-Shop Scheduling Problem},
pdfsubject={cs.IA},
pdfauthor={Imanol Echeverria},
pdfkeywords={Leveraging Constraint Programming in a Deep Learning Approach for Dynamically Solving the Flexible Job-Shop Scheduling Problem},
}

\begin{document}
\maketitle

\begin{abstract}
 Recent advancements in the flexible job-shop scheduling problem (FJSSP) are primarily based on deep reinforcement learning (DRL) due to its ability to generate high-quality, real-time solutions. However, DRL approaches often fail to fully harness the strengths of existing techniques such as exact methods or constraint programming (CP), which can excel at finding optimal or near-optimal solutions for smaller instances. This paper aims to integrate CP within a deep learning (DL) based methodology, leveraging the benefits of both. In this paper, we introduce a method that involves training a DL model using optimal solutions generated by CP, ensuring the model learns from high-quality data, thereby eliminating the need for the extensive exploration typical in DRL and enhancing overall performance. Further, we integrate CP into our DL framework to jointly construct solutions, utilizing DL for the initial complex stages and transitioning to CP for optimal resolution as the problem is simplified. Our hybrid approach has been extensively tested on three public FJSSP benchmarks, demonstrating superior performance over five state-of-the-art DRL approaches and a widely-used CP solver. Additionally, with the objective of exploring the application to other combinatorial optimization problems, promising preliminary results are presented on applying our hybrid approach to the traveling salesman problem, combining an exact method with a well-known DRL method.
\end{abstract}

\begin{multicols*}{2}
\raggedcolumns

\section{Introduction}\label{sec:intro}
The efficient resolution of the flexible job-shop scheduling problem (FJSSP) is key to operational success in numerous industrial and service-oriented sectors \citep{homayouni2021production, paschos2014applications}. Notably, its real-time resolution becomes increasingly important in rapidly changing environments \citep{cai2023real}. Solving the FJSSP differs from addressing the job-shop scheduling problem (JSSP) in that it involves tackling two distinct subproblems: one focused on the sequence of operations and the other on the allocation of machines. This complexity renders the FJSSP an NP-Hard problem \citep{garey1976complexity}, posing substantial challenges in devising scalable solutions.

Traditional approaches to addressing the FJSSP largely rely on exact methods, including mixed-integer linear programming and branch-and-bound algorithms \citep{xie2019review}. These methods frequently face significant computational complexity challenges, particularly in medium to large-sized problem instances. On the other end of the spectrum, heuristic methods, including various dispatching rules, can produce real-time solutions but typically fall short in quality \citep{garrido2000heuristic}. Meta-heuristic approaches, such as genetic algorithms, tabu search, and simulated annealing \citep{blum2003metaheuristics}, provide more robust solutions by iterating through numerous possibilities, but these methods too can be complex and computationally intensive, posing challenges for real-time application.

Recent advancements have increasingly favored machine learning, particularly deep reinforcement learning (DRL), for solving combinatorial optimization (CO) and scheduling problems \citep{mazyavkina2021reinforcement, song2022flexible}, aiming to achieve real-time solutions. DRL involves training a model (agent) to make a sequence of decisions. In scheduling, the agent learns to allocate resources and schedule operations by interacting with a simulated environment. It receives feedback in the form of rewards or penalties based on the efficiency of its decisions. Over time, the agent refines its strategy to optimize scheduling outcomes, aiming for time efficiency and solution quality. Another alternative for model training is the use of behavioral cloning (BC), where the model learns to imitate how to construct optimal solutions. This approach has been less explored due to the challenge of generating an optimal solution dataset for some CO problems \citep{bengio2021machine}.

The application of DRL in solving scheduling problems encounters several shortcomings. Firstly, the vast solution space makes finding optimal solutions through RL computationally intensive. This complexity may prevent the agent from training on such instances, thus limiting its effectiveness. Moreover, for smaller problem instances, exact methods like constraint programming (CP) can provide optimal solutions close to real-time \citep{da2019google}, but an RL-based approach cannot effectively incorporate such knowledge. Secondly, DRL methodologies often fail to utilize the full potential of established techniques in the field, such as exact methods and CP. This results in a significant missed opportunity to enhance performance by combining the strengths of both traditional and modern approaches.

In this paper, we focus on leveraging exact methods to enhance a DL-based methodology for the FJSSP. The main contributions are as follows:
\begin{itemize}
    \item We propose a methodology for training a DL model based on heterogeneous graph neural networks, using optimal solutions generated with CP. This allows for training a model with fewer instances and to achieve better results than methods trained using exclusively DRL.
    \item A new approach is introduced to generate solutions for FJSSP instances in real-time, combining the DL method with CP. Initially, DL is used to solve the instance until reaches a predefined complexity threshold, and then CP is employed for solving a reduced version of the instance, creating a superior combined approach.
    \item The effectiveness of our method is evaluated on three public FJSSP benchmarks, where it is compared with five state-of-the-art DRL-based methods and a widely-used CP method, OR-Tools. Our hybrid approach not only substantially outperforms these methods but also achieves competitive results with meta-heuristic algorithms, significantly reducing the time required to generate solutions.
    \item With the objective of exploring the application to other CO problems, preliminary results are presented, extending this hybrid approach to the traveling salesman problem (TSP), combining an exact method with a well-known DRL method, thereby improving its performance.
\end{itemize}  

The structure of the paper is as follows: A review of DL methods in scheduling is presented in Section \ref{sec:related}. The formulation of FJSSP is detailed in Section \ref{sec:formulation}. Our proposed method is elaborated in Section \ref{sec:method}. Section \ref{sec:experiments} is devoted to discussing the experiments and results. Finally, the paper presents conclusions and a discussion on future work in Section \ref{sec:conclusion}.

\section{Related work}\label{sec:related}
In this section, we examine previous applications of DL methods to CO problems. Initially, we discuss the application of DL to routing problems such as the TSP and the capacitated vehicle routing problem (CVRP), which lay the groundwork for scheduling problems. We then turn our attention to scheduling, highlighting DL's application in this area and reviewing some hybrid approaches.

\begin{table*}[t]
\caption{Survey of DL-methods for solving scheduling problems in real-time.}
\label{tab:related}
\resizebox{\textwidth}{!}{
\begin{tabular}{l r r r r r}
        \hline
        Method &Problem & Learning method & Feature extractor & Size-agnostic & Hybrid approach  \\
        \hline
        \hline
         \cite{lei2022multi} & FJSSP & RL& GIN &Yes& No\\
         \cite{zhang2023deepmag} &FJSSP&RL&MLP&Yes&No\\
         \cite{song2022flexible} &FJSSP&RL& HGAT &Yes& No \\
         \cite{ho2024residual} & JSSP, FJSSP & RL& HGAT&Yes&No \\
         \cite{wang2023flexible} & FJSSP&RL&HGAT&Yes&No\\
         \cite{echeverria2023solving} &FJSSP&RL&HGAT&Yes&No\\
         \cite{yuan2023solving}&FJSSP&RL&MLP&Yes&No\\
         \cite{zhang2020learning}&JSSP&RL&GNN&Yes&No\\
         \cite{tassel2023end}&JSSP&RL&Custom NN&Yes&CP feedback\\
         \cite{ingimundardottir2018discovering}&JSSP&BC&MLP&No&No\\
         \cite{li2022learning}&FSS&BC&GNN&Yes&No\\
         \cite{chen2023two}&FJSSP&BC&MLP&No&No\\
         \textbf{Ours}&FJSSP&BC&HGAT&Yes&Jointly created solutions\\
         \hline
    \end{tabular}}
\end{table*}

For routing problems, one of the earliest approaches was proposed by \cite{vinyals2015pointer}, introducing the Pointer Network (Ptr-Net) for multiple CO problems, including the TSP, training the network with optimal solutions. \cite{bello2016neural} extended this work using DRL, particularly the REINFORCE algorithm \citep{williams1992simple}, to train the Ptr-Net for solving the TSP, and \cite{nazari2018reinforcement} extended it to the CVRP. The application of attention models to CO problems by \cite{kool2018attention} was a significant advancement, replacing the Ptr-Net with a transformer \citep{vaswani2017attention}. Thanks to transformers, it is possible to better capture the relationships between the different elements of a CO problem, leading to improved performance in solving these complex tasks. The policy optimization for multiple optima (POMO) method, proposed by \cite{kwon2020pomo}, achieved significant improvements by exploiting symmetries in the TSP and CVRP. POMO improved the sampling method, which consists of generating a set of solutions instead of one, based on a new augmentation-based inference method.

More recently, the application of DL, mainly DRL, has been extended to scheduling problems. In comparison to problems like the TSP or CVRP, scheduling problems have the additional complexity of having different elements (operations, jobs, and machines), along with different types of relationships between them. This implies the need for neural networks (NNs) to adequately extract information from an instance to find an optimal policy. Additionally, it is crucial that these NNs to be size-agnostic, meaning they should be capable of solving instances of varying sizes to avoid the need for training numerous networks for different instance sizes. In this context, building on the presented work in routing problems, \cite{zhang2020learning} modelled the JSSP using the disjunctive graph representation and proposed an end-to-end DRL method to learn dispatching rules. The use of graphs and graph neural networks (GNNs) addresses the challenges of size-agnosticism and effectively captures the complete range of information required for these complex scheduling problems.

Expanding on this work to the FJSSP, using this representation, \cite{lei2022multi} approached the problem as multiple Markov
decision processes (MDPs), employing multi-pointer graph networks, specifically a graph isomorphism network (GIN), and proximal policy optimization as the RL algorithm. In this approach, two policies are learned and applied sequentially: one for operation selection and the second for machine selection. \cite{song2022flexible} proposed the application of heterogeneous graph attention networks (HGATs), creating nodes for operation and machine types with various types of edges indicating precedence between operations and the machines on which each operation can be performed. Compared to the disjunctive graph representation, fewer edges are needed, simplifying the action space. Multiple papers have delved deeper into the use of heterogeneous graphs, proposing various enhancements. \cite{ho2024residual} introduced a new way to model the JSSP and FJSSP, removing irrelevant machines and operations, such as completed jobs, to facilitate policy decision-making. \cite{wang2023flexible} leveraged attention models for deep feature extraction and proposed a dual-attention network that outperformed various benchmarks in the literature. \cite{echeverria2023solving} introduced a new modelling approach for the FJSSP, using job types as nodes and two strategies to improve the results of their DRL-based methodology: the use of DRs to limit the action space and a strategy to generate diverse policies. \cite{yuan2023solving}, instead of using a GNNs, which can be computationally expensive, proposed a lightweight multi-layer perceptron (MLP) as the state embedding network. Lastly, \cite{zhang2023deepmag} proposed a multi-agent approach that employed deep Q-Networks, in conjunction with multi-agent graphs to address the FJSSP. In this model, each machine and job are treated as an individual agent, and these agents cooperate based on the multi-agent graph, representing operational relationships between machines and jobs.

In scheduling, various approaches have used optimally generated solutions to train a model with BC to address these problems, which can improve the network's capability as the training set is of higher quality. This contrasts with RL, where the training process necessitates the exploration of sub-optimal experiences, which is time-consuming, to discover optimal solutions. \cite{ingimundardottir2018discovering} proposed a method to learn an effective dispatching rule for the JSSP using a mixed-integer programming solver to generate optimal samples. \cite{li2022learning} tackled the flow shop scheduling (FSS), a simpler variant of the JSSP, using GNNs to generate a policy that imitates a sequence of actions and states generated by the NEH algorithm \citep{sharma2021improved}, improving upon an NN trained through RL. The authors of \cite{chen2023two} propose a method to solve the FJSSP by splitting it into two problems: an operation-to-machine assignment problem and a sequencing problem, for which they train separate convolutional networks with instances solved by the IBM CP-Optimizer solver \citep{laborie2009ibm}. However, by using this type of network, their approach is not size-agnostic, and they train a model for each problem instance, limiting its applicability. 

Finally, there are approaches that hybridize machine learning and classic methods. Machine learning-based methods can be used to configure another algorithm based on instance information extracted by the NN or alongside optimization algorithms. An example of the former is the work of \cite{muller2022algorithm}, where they train a model that, based on instance features, predicts which of five CP solvers is most suitable. Related to the latter, for example, many mixed-integer linear programming methods use the branch-and-bound algorithm, and \cite{lodi2017learning} studied the capabilities of machine learning models in selecting the branching variable. These approaches demonstrate how machine learning can enhance exact methods but do not focus on generating solutions dynamically. Lastly, \cite{tassel2023end} leveraged existing CP solvers to train a network with RL based on the generic CP encoding of the JSSP and receiving feedback from the CP method, achieving better results than static priority dispatching rules and even outperforming a CP solver, Choco \citep{Prud'homme2022}.

Table \ref{tab:related} presents a summary of recent work in solving scheduling problems in real-time, with a special focus on the FJSSP. It reveals that most current approaches primarily rely on RL for real-time FJSSP resolution, which implies that they cannot incorporate optimal solutions. The use of BC to scheduling problems, in contrast, is less prevalent. Furthermore, the hybridization of methods — combining different approaches to enhance solution quality — remains an under-explored strategy in this field.

\section{FJSSP formulation}\label{sec:formulation}


An instance of the FJSSP is characterized by a set of jobs $\mathcal{J} = \{j_1 , j_2, \ldots, j_n \}$, where each $j_i \in \mathcal{J}$ comprises a set of operations $\mathcal{O}_i = \{ o_{i1}, o_{i2}, \ldots o_{im}\}$, and each operation can be executed on one or more machines, from the set $\mathcal{M} = \{m_1, m_2, \ldots , m_p\}$. The processing time of operation $o_{ij}$ on machine $m_{k}$ is denoted as $p_{ijk} \in \mathbb{R}^+$. We define $\mathcal{M}_{o_{ij}} \subseteq \mathcal{M}$ as the subset of machines on which operation $o_{ij}$ can be processed, $\mathcal{O}_{j_i} \subseteq \mathcal{O}$ as the set of operations belonging to job $j_i$ where $\bigcup\limits_{i = 1}^n \mathcal{O}_{j_i} = \mathcal{O}$, and $\mathcal{O}_{m_k} \subseteq \mathcal{O}$ as the set of operations that can be performed on machine $m_k$.

\begin{figure*}[!t]
  \centering
  \begin{minipage}[!t]{0.48\textwidth}
    \includegraphics[width=1\textwidth]{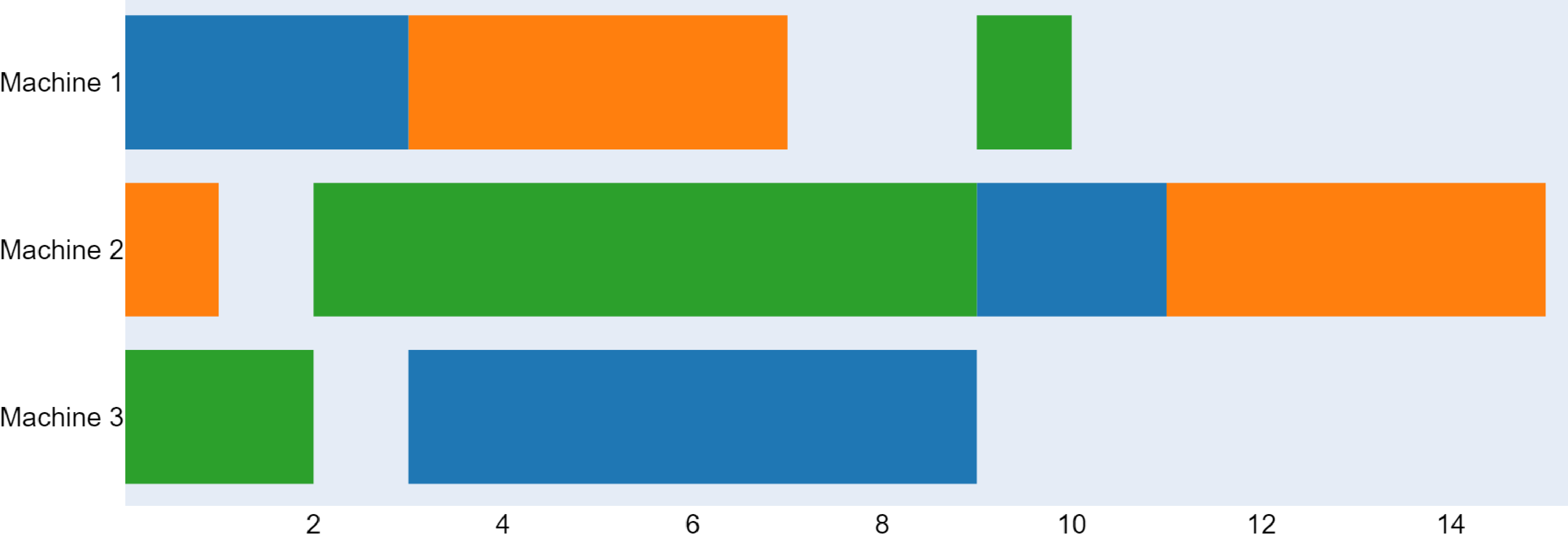}
    \caption{A solution to the FJSSP instance with a makespan of 16.}
    \label{fig:sol}
  \end{minipage}
  \begin{minipage}[!t]{0.48\textwidth}
    \includegraphics[width=1\textwidth]{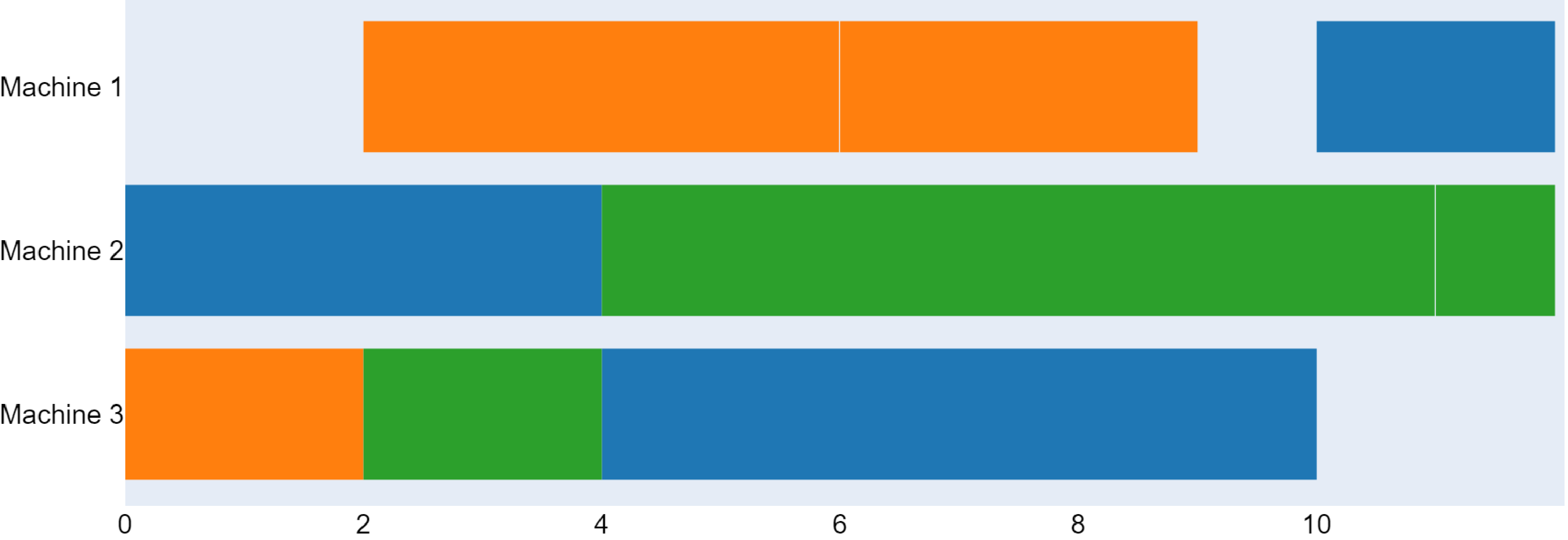}
    \caption{An optimal solution with a makespan of 12.}
    \label{fig:optsol}
  \end{minipage}
\end{figure*}

The assignment of operations on machines must adhere to a set of constraints:

\begin{itemize}
\item All machines and jobs are available at time zero.
\item Each machine can execute only one operation at a time, and the execution cannot be interrupted.
\item An operation can be performed on only one machine at a time.
\item The execution order of the set of operations $\mathcal{O}_i$ for every $j_i \in \mathcal{J}$ must be respected.
\item Job executions are independent of each other, meaning no operation from any job precedes or takes priority over the operation of another job.
\end{itemize}

The FJSSP integrates two problems: a machine selection problem, in which the optimal machine is selected for each operation, and a routing and sequencing problem, where the order of operations on a machine needs to be determined. Given an assignment of operations to machines, the completion time of a job, $j_i$, is defined as $C_{j_i}$, and the makespan of a schedule is defined as $C_{\text{max}} = \max\limits_{j_i \in \mathcal{J}} C_{j_i}$, which is the most common objective function to minimize.

In Table \ref{tab:fjssp_example}, an example of a small FJSSP instance is presented. In this instance, there are three machines and three jobs, each with three operations that need to be processed. The table indicates which machines can perform which operations and their corresponding processing times. Figures \ref{fig:sol} and \ref{fig:optsol} present two solutions, with the latter being optimal. The x-axis represents the temporal axis, and the y-axis represents the machines where operations need to be assigned. The coloured bars represent the different operations of the jobs (the same colour represents the same job), and each operation must wait until the preceding operation is completed, except for the first operation of that job. In the optimal solution, it can be observed how the jobs are more executed in parallel, reducing the number of gaps between operations and minimizing the makespan.

\begin{table}[H]
\caption{FJSSP instance with 3 jobs, 3 machines, and 9 operations.}
\label{tab:fjssp_example}
\begin{tabularx}{\columnwidth}{X X X X X X}
\toprule
Jobs  & Operations & \(m_1\) & \(m_2\) & \(m_3\) \\ \midrule
\(j_1\) & \(o_{11}\) & 3 & 4 & - \\
       & \(o_{12}\) & - & 5 & 6 \\
       & \(o_{13}\) & 2 & - & 3 \\
\hline
\(j_2\) & \(o_{21}\) & - & 1 & 2 \\
       & \(o_{22}\) & 4 & - & 5 \\
       & \(o_{22}\) & 3 & 4 & - \\
\hline
\(j_3\) & \(o_{31}\) & - & - & 2 \\
       & \(o_{32}\) & 6 & 7 & - \\
       & \(o_{33}\) & 1 & 1 & 2 \\ \bottomrule
\end{tabularx}
\end{table}

The FJSSP can be modelled as a CP as described in Model \ref{alg:sfjsspcp}. To do this, two decision interval variables are defined, $o_{ij}$ for each operation $o_{ij} \in \mathcal{M}$, and for each option that the operation has $m_k \in \mathcal{M}{ij}$, an optional decision variable $mo_{ijk}$ is defined. The objective function (\ref{sfjsspcp:MP1}) consists of minimizing the makespan. Constraint (\ref{sfjsspcp:MP2})  is to assure that an operation cannot start until its precedent is completed, Constraint (\ref{sfjsspcp:MP3}) specifies that an operation $o_{ij}$ can only be assigned to a compatible machine among all its alternatives $[ mo_{ijk}]_{\forall m_k \in M_{ij}}$, and Constraint (\ref{sfjsspcp:MP4}) indicates that a machine can process only one operation at a time, that is, if the optional decision variable, $mo_{ijk}$ is selected, no other operation can be assigned to that machine at the same time.

\begin{algorithm}[H]
\small
\floatname{algorithm}{Model}
\caption{FJSSP CP model}\label{alg:sfjsspcp} 
\vspace{-4mm}
\begin{align}
    &\text{min} \qquad && \max\limits_{o_{ij} \in \mathcal{O}} \texttt{EndOf}(o_{ij}) \label{sfjsspcp:MP1} \\ 
   & \text{s.t.} \qquad && \texttt{EndBeforeStart}(o_{ij}, o_{ij+1}) \nonumber \\ 
    &&& \quad \forall o_{ij} \in j_i \, |\, j > 0 , \forall J_i \in \mathcal{J}  \label{sfjsspcp:MP2} \\
    &&& \texttt{Alternative}(o_{ij}, [ mo_{ijk}]_{\forall m_k \in M_{ij}}) && \forall o \in \mathcal{O} \label{sfjsspcp:MP3} \\
    &&& \texttt{NoOverlap}(mo_{ijk}) && \forall m_k \in \mathcal{M}\label{sfjsspcp:MP4}
\end{align}
\vspace{-5mm}
\end{algorithm}

\section{Proposed method: BCxCP}\label{sec:method}
\begin{figure*}[!t]
    \centering
    \includegraphics[width=\textwidth]{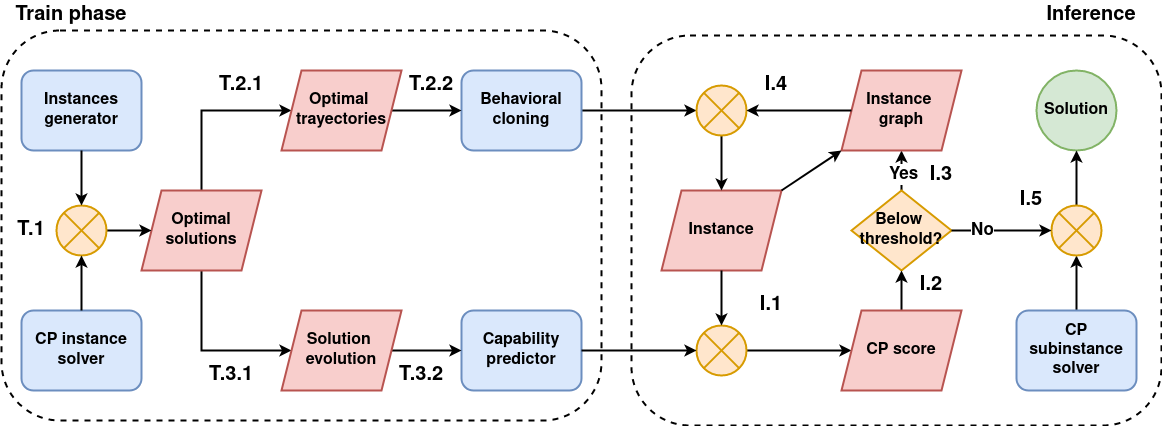}
    \caption{Diagram depicting the components of the methodology and their interaction during the training and inference phases.}
    \label{fig:diagram}
\end{figure*}

In this section, we introduce our approach for the real-time resolution of FJSSP instances, integrating CP into a DL-based framework. This hybridization, which we refer to as BCxCP, leverages the ability of CP methods to optimally solve small instances. Our methodology, depicted in Figure \ref{fig:diagram}, utilizes the advantages of CP in two distinct manners.

\begin{itemize}
\item First, we train an HGAT with optimal solutions solved by CP (T.1). For this, the FJSSP is formulated as a Markov process, requiring the definition of state and action spaces and a transition function. This paper employs heterogeneous graphs to represent an instance, as it effectively captures the elements of the problem (jobs, operations and machines) and their relations. The action space involves assigning operations to machines. The solution generated by CP is transformed into a trajectory of states and actions, delineating the optimal action for each state (T.2.1). A model is then trained via BC (T.2.2) to emulate this trajectory, sequentially assigning operations to machines until none remains.
\item Second, we employ CP to jointly generate solutions. Addressing the uncertainty of when an instance can be solved by CP —a larger instance might be easier to solve due to a reduced number of machine assignment options for the operations— we propose a supervised learning method named the CP capability predictor. This method assesses whether an instance is solvable by CP in real time (T.3.1), using the evolution of solution quality over time based on the instance's characteristics to refine the model (T.3.2). 'Solution quality over time' refers to predicting the potential for improvement from real-time to a longer term, where minimal improvement indicates an optimal or pseudo-optimal result.
\end{itemize}

Therefore, during inference, when a new instance (I.1) needs to be solved, the CP capability predictor is applied to obtain a score (I.2), ranging from 0 to 1 - with 1 indicating certainty that it can be solved by the CP method in real-time. If this confidence level is not higher than a threshold, it is transformed into a graph (I.3) and the BC method is applied by assigning an operation to a machine (I.4), reducing the size and complexity of the problem. The process of assigning a score is repeated until it exceeds the threshold, and the remaining part of the instance is solved using a modified model of the CP method that already has a set of operations assigned to machines.

The following subsections will detail the modeling of the problem as a Markov process, the definition of the HGAT for extracting information from a graph, and elaborate on the CP capability predictor and the hybridization strategy.

\subsection{The FJSSP as a Markov Process}
The first step in our methodology involves modeling the problem as a Markov process. The key distinction between Markov processes and MDPs lies in the requirement of a reward function for MDPs. Since a method based on BC does not need this, defining an Markov process only requires specifying a state space, an action space, and a transition function.

\textbf{State space.} This paper adopts and extends the state space representation based on heterogeneous graphs previously presented \citep{echeverria2023solving} (section 4.2, pages 7-8). These graphs incorporate different node types, edge types, and features. At each step $t$, the state $s_t$ is defined as a heterogeneous graph $\mathcal{G}_t=(\mathcal{V}_t, \mathcal{E}_t)$, where the set $\mathcal{V}_t$ consists of three types of nodes (operations, jobs, and machines), and $\mathcal{E}_t$ comprises six types of edges, both directed and undirected. The direction of the edges is crucial as node information flows according to their direction, as it is explained in the next section on the GNN architecture. Detailed information about all node features is provided in \ref{appendix:nodefatures}. The edges in the graph are characterized as follows:
\begin{itemize}
\item Undirected edges between machines and operations. This relationship indicates the machines where an operation can be executed. An undirected edge is represented as two directed edges between machines and operations and vice versa. Information about this relationship, such as its processing time and other features, is added to this undirected edge.
\item Directed edges between operations and jobs to represent the operations that belong to a job.
\item Directed edges between operations indicating that one operation immediately precedes another.
\item Directed edges between machines and jobs. Each job is linked to machines capable of executing its first operation. If all operations have already been performed, this node will not be connected to any machine. Similar to the edges between operations and machines, this edge also has features, such as its processing time.
\item Jobs are connected to each other because the state of a job and the operations it has yet to perform will influence the importance of completing the other jobs.
\item Machines are connected to each other for a similar reason to jobs, as they condition each other.
\end{itemize}

\textbf{Action space.} At each timestep $t$, the action space $\mathcal{A}_t$ corresponds to compatible jobs and machines. When selecting a job, the first operation of that job which has not been scheduled yet is chosen. This decision-making process can be modeled as an edge classification problem, identifying the most suitable pairing. \\

\textbf{Transition function.} The solution is constructed by sequentially assigning jobs to machines. When an operation is assigned, it is removed, and the edges of the selected job are updated to reflect the next operation to be executed, as well as the features of all the nodes.

\subsection{Graph Neural Network Architecture}

The use of GNNs is common for extracting features from a heterogeneous graph, particularly those using attention mechanisms \citep{song2022flexible, wang2023flexible}. In this paper, we adapt the architecture proposed by \cite{shi2020masked} and \cite{hu2020heterogeneous}.

We will only show how to calculate the embedding of a machine node, but the other nodes are computed in a similar fashion. A machine node is related to the operations that can process, all the other machines, in order to share information about their respective state to establish a priority, and with the jobs that can immediately schedule an operation in a machine. The initial representation of a machine $m_k$ is named $\boldsymbol{h_{m_{k}}} \in \mathbb{R}^{d_{\mathcal{M}}}$, where $d_{\mathcal{M}}$ is the initial number of machine features. Analogously, for an operation $o_{ij}$ its initial representation is $\boldsymbol{h_{o_{ij}}}$ and the edge connecting both nodes is denoted $\boldsymbol{h_{mo_{kij}}} \in \mathbb{R}^{d_{\mathcal{OM}}}$, where $d_{\mathcal{OM}}$ is the dimension of the feature. The first step consist of calculating the attention coefficients for the three relationships. The coefficient between the machine and an operation, $o_{ij}$, is computed as:
\begin{equation}
\alpha_{mo_{kij}}=\operatorname{a}\left(\frac{\left(\mathbf{W}_1^{\mathcal{MO}} \boldsymbol{h_{m_{k}}}\right)^{\top}\left(\mathbf{W}_2^{\mathcal{MO}} \boldsymbol{h_{o_{ij}}} + \mathbf{W}_3^{\mathcal{MO}} \boldsymbol{h_{mo_{kij}}}\right )}{\sqrt{d_{\mathcal{MO}}^\prime}}\right)
\end{equation}
where $\operatorname{a}$ is an activation function (in our case a softmax), $\boldsymbol{W_1^{\mathcal{MO}}} \in \mathcal{R}^{d_{\mathcal{MO}}^\prime \times d_{\mathcal{MO}}}$, $\boldsymbol{W_2^{\mathcal{MO}}} \in \mathcal{R}^{d_{\mathcal{MO}}^\prime \times d_{\mathcal{O}}}$, and $\boldsymbol{W_3^{\mathcal{MO}}} \in \mathcal{R}^{d_{\mathcal{MO}}^\prime \times d_{\mathcal{MO}}}$ are learned linear transformations, and $d_{\mathcal{MO}}^\prime$ is the dimension given to the hidden space of the embeddings. The next two attention coefficients, $\alpha_{mm_{kk^\prime}}$ between the machines $m_k$ and $m_{k^\prime}$, and $\alpha_{mj_{ki}}$ between the machine $m_k$ and the job $j_{i}$ have been computed in a similar way.  

Once the attention coefficients have been calculated, the new embedding of $m_{k}$, $\boldsymbol{h_{m_{k}}^\prime}$ is computed as:
\begin{align}
\boldsymbol{h_{m_{k}}^\prime}= \mathbf{W}_1^{\mathcal{M}}\boldsymbol{h_{m_{k}}} + \sum_{m_k^\prime \in \mathcal{M} \; | \; k != k^\prime} \alpha_{mm_{kk^\prime}} \mathbf{W}_2^{\mathcal{M}} \boldsymbol{h_{m_{k^prime}}} + \nonumber \\
\sum_{o_{ij} \in \mathcal{O}_{m_{k}}} \alpha_{mo_{kij}} \left(\mathbf{W}_4^{\mathcal{MO}} \boldsymbol{h_{o_{ij}}} + \mathbf{W}_5^{\mathcal{MO}} \boldsymbol{h_{mo_{ijk}}}\right) + \nonumber \\
\sum_{j_i \in \mathcal{JM}_{m_{k}}} \alpha_{mj_{ki}} \left(\mathbf{W}_1^{\mathcal{JM}} \boldsymbol{h_{j_{i}}} + \mathbf{W}_2^{\mathcal{JM}} \boldsymbol{h_{mj_{ki}}}\right) 
\end{align}
where $\mathcal{JM}_{m_k}$ is the set of edges connecting compatible jobs with the machine $m_k$, $\boldsymbol{h_{j_{i}}}$ is the embedding of the job $j_i$ and $\boldsymbol{h_{mj_{ki}}}$ are the features of the edge connecting both.

To enhance the GNN's ability to capture more meaningful relationships between nodes, two strategies can be employed: using multiple attention heads and increasing the number of layers. The utilization of multiple attention heads is a widely adopted technique in diverse domains such as neural machine translation, computer vision \cite{chen2015abc, behnke2020losing}, and in the context of CO problems \cite{nazari2018reinforcement, wang2023flexible}.

Assuming $K$ attention heads, $K$ different embeddings are computed, denoted as $\boldsymbol{h_{o_{ij}}^{\prime k}}$ for $k = 1,2, \ldots K$. Each attention head possesses its unique attention coefficients along with corresponding linear transformations. Subsequently, these  embeddings are concatenated to yield the final embedding, represented by the equation:
\begin{equation}
\boldsymbol{h_{o_{ij}}^\prime}=\|_{k=1}^K \boldsymbol{h_{o_{ij}}^{\prime k}}
\end{equation}

The second avenue for augmenting complexity involves employing multiple layers. This entails repeating the entire procedure $L$ times, each time employing different learnable parameters. A sufficient number of layers is crucial to facilitate the propagation of information from nodes not directly connected. In each layer $l$, embeddings calculated in the preceding layer $l-1$ are utilized. The output of this layer is denoted as $\boldsymbol{h_{o_{ij}}^L}$.

\subsection{Training procedure with BC}

Training the agent using BC requires generating a set of $n_i$ optimal solutions, $\mathcal{S}$. To do this, a set of instances that can be solved by a CP method in less than 60 seconds is synthetically generated, and then a trajectory of states and actions is built. Since the BC approach is constructive, feasible assignments of operations to machines must be established. To achieve this, operations are first sorted by their start time and then by their end time. This order allows the construction of a sequence of states and actions. Formally, for each solution $e \in \mathcal{S}$, a trajectory $\mathcal{T}_{e} = \{ (s_1, a_1), (s_2, a_2), \ldots, (s_{|\mathcal{O}|}, a_{|\mathcal{O}|}) \}$ is created, where $\mathcal{O}$ is the set of operations of the instance being solved. The number of actions corresponds to the number of operations, as an action consists of assigning an operation to a machine. For all $i = \{1,2, \ldots |\mathcal{O}| \}$, $s_i$ is a heterogeneous graph, and $a_i$ is the action to be taken, which is a vector with all elements equal to 0 except for a single element indicating the taken action. Lastly, the dataset is defined as the union of all the trajectories, $\mathcal{D}_{BC} = \bigcup\limits_{e \in \mathcal{S}} \mathcal{T}_{e}$.

The set of trajectories constitutes the dataset used to train the agent as a supervised learning model, specifically as a classification problem. This is done over $n_e$ epochs, with the dataset divided into batches of size $n_b$ for each epoch. The next step is to define a loss function. In BC and other classification problems, cross-entropy loss is widely used. However, this paper proposes using the KL-Divergence loss because preliminary experiments showed more stability, as suggested by \cite{czarnecki2019distilling} for policy distillation. 

Finally, to generate a policy, $\pi_\theta(a_t | s_t)$, with the embeddings calculated by the GNN, MLPs are used to define the policy on edges connecting compatible jobs and machines. If, for the state $s_t$, this set of edges is denoted as $\mathcal{MJ}_t$, the probability of selecting an edge $e \in \mathcal{MJ}_t$, where this edge connects job $j_i$ and machine $m_k$, and the edge has features $\boldsymbol{h_{mj_{ki}}}$, is defined as:

\begin{equation}
\mu_\theta(e | s_t) = \text{MLP}_{\theta}(\boldsymbol{h_{j_{i}}^L} || \boldsymbol{h_{m_{k}}^L} || \boldsymbol{h_{mj_{ki}}} )
\end{equation}
\begin{equation}
\pi_\theta(e | s_t) = \frac{\mu_\theta(e | s_t)}{\sum_{x \in \mathcal{MJ}_t }{\mu_\theta(x | s_t)}}
\end{equation}

Finally, for a policy $\pi_\theta$, the loss function is calculated as:

\begin{equation}
\sum _{(s_i, a_i) \in {\mathcal {D}_{BC}}} a_i \log \left({\frac {a_i}{\pi_\theta(s_i)}}\right)\end{equation}

where $\pi_\theta(s_i)$ is the probability distribution over all actions.

\subsection{CP Capability Predictor} \label{ss:predictcomplex}

As mentioned in the introduction, besides training a method based on instances solved by CP, another way to leverage a CP method is for jointly constructing solutions. Initially, the method trained via BC is used, assigning operations until the instance becomes sufficiently simple. This necessitates the ability to predict when an instance is considered simple enough for a CP method to resolve in real-time. To address this, we frame the issue as a supervised learning problem and name the resulting method the CP capability predictor. There is no precise definition of real-time in existing literature; hence, for the purposes of this paper, we consider a method to be real-time if it can generate a solution within $|\mathcal{O}|*om$ seconds for an instance with $\mathcal{O}$ operations, where $om$ is a parameter assumed to be close to 0. In our paper, we adopted a value of 0.01. This value makes the method comparable to the execution time of other methods based on DRL, and it aims to balance speed with solution quality.

The first step to solve this problem is to create a dataset comprising instances of varying sizes and solving them while saving how the solution quality evolved over time. It is crucial for these instances to differ not only in the number of machines, jobs, and operations but also in their processing times and the range of machines an operation can utilize, ensuring a varied dataset. For each solution in the dataset $e \in \mathcal{S}_{CP}$, a sequence of solutions is generated $\mathcal{S}_e = \{ (C_1^e, t_1), (C_2^e, t_2), \ldots, (C_m^e, t_m) \}$, where $m$ is the number of solutions found for the instance, $C_i^e$ represents the makespan of the i-th solution obtained at time $t_i$. This allows us to obtain the variable to predict. Given that $C_j^e$ is the last makespan such that $t_j \leq d_{n_\mathcal{O}}*0.01$, the variable to predict will be $\frac{C_j^e}{C_m^e}$. The minimum value of this variable will be 0, indicating that the CP method failed to produce any result in less than $d_{n_\mathcal{O}}*0.01$ time, suggesting that the CP method is not suitable for that instance. The maximum value will be 1, indicating that the CP method did not improve beyond what was found in a short time, suggesting that the solution found quickly is optimal or near-optimal.

\begin{figure}[H]
\includegraphics[width=0.5\textwidth]{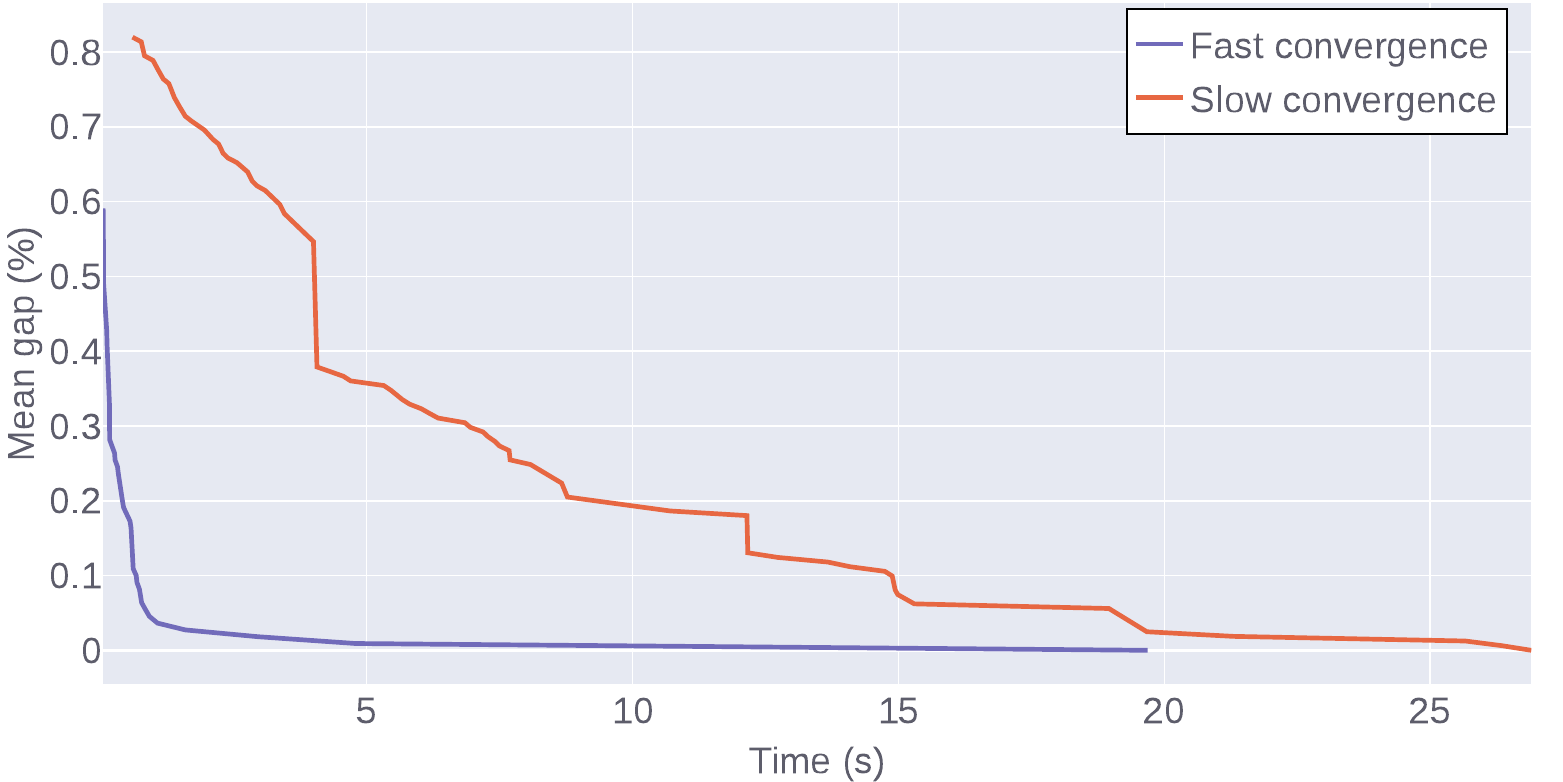} 
\caption{The evolution of solution quality over time is depicted for two instances} 
\label{fig:differentsolutions} 
\end{figure}

In Figure \ref{fig:differentsolutions}, two examples of these solution types are illustrated, with the x-axis representing time and the y-axis showing the gap relative to the last obtained solution, which may not necessarily be optimal. As observed, the blue instance quickly reaches a near-optimal solution (halted as it found the optimum in less than 30 seconds). However, the red instance converges much more slowly, indicating potential room for improvement.

With the predictive variable established, we frame the issue as a supervised learning regression problem, necessitating the specification of a prediction algorithm and the identification of predictor variables. It is crucial for this methodology to be size-agnostic, ensuring it remains effective regardless of the instance's scale, and to maintain efficiency to avoid excessive computational overhead. To satisfy these criteria, we select predictor variables based on instance characteristics and opt for a conventional regression model, such as a gradient boosting regressor, over more complex alternatives. Table \ref{tab:cpvariable} presents the name and a description of the predictive variables used in the model. Subsection \ref{exp:cpcap} will explore the significance of these variables for the effectiveness of this method.

\begin{table}[H]
\small
\centering
\caption{Predictive variables used for the CP capability predictor.}
\label{tab:cpvariable}
\begin{tabularx}{\linewidth}{p{2.5cm} X}
Name & Description \\
\midrule
N. operations & Number of operations of the instance. \\
N. jobs &  Number of jobs of the instance. \\
N. machines &  Number of machines of the instance. \\
Mean options & Mean of the number of operations a machine can process. \\
Std. options & Standard deviation of the number of operations a machine can process. \\
Skewness options &  Skewness of the number of operations a machine can process. \\
Min pt. & Minimum processing time of the instance.\\
Max pt. & Maximum processing time of the instance. \\
Span pt. & Difference between the maximum and minimum processing times.\\
\bottomrule

\end{tabularx}
\end{table}

\subsection{Jointly constructing solutions}

Once defined how to generate an agent based on BC and a method to predict the ability of an CP method to solve an instance in real-time, we will now propose a way to hybridize them with the aim of obtaining a better result than either of its components. The main idea is to keep assigning operations to machines until the instance becomes simple enough; that is, until the model prediction exceeds a constant, which is similar but less than 1 (in this paper, it is considered to be 0.98), representing the estimated gap from the optimum. Although this approach suggests a method to hybridize it with an agent trained with BC, it could also be applied to an agent based on RL, as demonstrated in its application to the TSP, presented in Subsection \ref{sub:tsp}.

\begin{algorithm}[H]
\small
\floatname{algorithm}{Model}
\caption{FJSSP CP subinstance model}
\label{alg:sfjssspcp} 
\vspace{-4mm}
\begin{align}
    &\text{min} \qquad && \max\limits_{o_{ij} \in \mathcal{O}} \texttt{EndOf}(o_{ij}) \label{sfjssspcp:MP1} \\ 
    & \text{s.t.} \qquad && \texttt{EndBeforeStart}(o_{ij}, o_{ij+1}) \nonumber \\ 
    &&& \qquad \forall o_{ij} \in j_i \, |\, j > 0 , \forall J_i \in \mathcal{J}  \label{sfjssspcp:MP2} \\
    &&& \texttt{Alternative}(o_{ij}, [ mo_{ijk}]_{\forall m_k \in M_{ij}}) & \forall o \in \mathcal{O} \label{sfjssspcp:MP3} \\
    &&& \texttt{NoOverlap}(mo_{ijk}) & \forall m_k \in \mathcal{M}\label{sfjssspcp:MP4} \\
    &&& \texttt{StartOf}(o_{ij}) \geq sj_i  & \forall j_i \in \mathcal{J}\label{fjssspcp:MP5}\\
    &&& \texttt{StartOf}(mo_{ijk}) \geq sm_k  & \forall m_k \in \mathcal{M}\label{fjssspcp:MP6}
\end{align}
\vspace{-5mm}
\end{algorithm}

 To achieve this, first, we need to adapt the CP formulation for the standard FJSSP presented in Model \ref{alg:sfjssspcp}, to solve a subinstance of the FJSSP, i.e., an instance that already has some assigned operations. Two new constraints need to be added. These constraints are related to machines starting after their last assigned operation finishes and jobs starting after their last assigned operation. We define $sj_i$ for all $j_i \in \mathcal{J}$ as the minimum time at which all operations of $j_i$ must start, and $sm_k$ for all $m_k \in \mathcal{M}$ as the minimum instant at which an operation assigned to machine $m_k$ can start. These constraints are defined in (\ref{fjssspcp:MP5}) and (\ref{fjssspcp:MP6}), respectively.

\section{Experiments}\label{sec:experiments}
This section details the experiments to evaluate the effectiveness of our hybrid approach, BCxCP. We begin with the results for the FJSSP and then present preliminary findings for the TSP to assess the hybrid approach's applicability to other CO problems.

\subsection{Experimental setup}

In this subsection, we will present the experimental setup, explaining the method's configuration, the instance generation and benchmark dataset, and the evaluation metrics and baselines.

\subsubsection{Configuration}

The proposed method was implemented using Python 3.10, with Pytorch Geometric for the GNNs \citep{Fey/Lenssen/2019} and Scikit-learn \citep{scikit-learn} for the the CP capability predictor. OR-Tools \footnote{ We obtained the implementation from \url{https://github.com/google/or-tools/blob/stable/examples/python/flexible_job_shop_sat.py}.} was selected as a CP solver for its open-source availability and widespread use in scheduling problems \citep{da2019industrial, muller2022algorithm}. Experiments were run on hardware featuring an AMD Ryzen 5 5600X processor and an NVIDIA GeForce RTX 3070 Ti GPU. The code will be made available after the paper is accepted.

To find the best hyperparameters for the BC method, Bayesian optimization was used, employing the Tree-structured Parzen Estimator as the sampler. The optimized hyperparameters for the GNN include the number of GNN layers $L \in [1, 3]$, the number of hidden channels $d^\prime \in \{32, 64, 128\}$ of the GNN and the MLPs, the learning rate $lr \in [0.00001, 0.001]$, and the batch size $b \in \{32,64\}$. The maximum batch size, $n_b$, was set to 64, as a larger size was not feasible due to GPU capacity limitations, and the number of episodes, $n_e$, was set to 25. The final configuration was set to: $L=3$, $d^\prime = 128$, $n_b = 64$ and $lr=0.0002$.

\subsubsection{Instance generation and benchmark dataset}

Due to the necessity to train two models, the first step involves generating two distinct datasets. The same dataset was not used for both due to computational reasons (specifically GPU capacity); the BC model needed to be trained on smaller instances, whereas the CP capability predictor could be trained with instances of varying sizes.

To generate the dataset for the BC model, a total of 1000 instances were generated for training and 50 for validation. The method for generating instances is based on the one proposed by \cite{brandimarte1993routing}. For the CP Capability predictor, 1000 instances were generated using 200 as the validation set. In this case, the validation set was larger due to a lower computational cost for validation. For each instance, OR-Tools was used to solve it within a time limit of 1 minute. Table \ref{tab:instanceparams} shows the ranges of the parameters that were uniformly sampled to create the instances for both datasets. These include the number of jobs and machines, the number of operations for each job, the number of machines each operation can be processed on, the average processing time for each operation $\overline{p}$, and its deviation $d$. Based on these last two values, the processing time for each operation option is sampled from a uniform distribution $\mathcal{U}(\overline{p}(1-d), \overline{p}(1+d))$.

\begin{table}[H]
\small
\centering
\caption{Ranges of the parameters employed to generate instances for both datasets.}
\label{tab:instanceparams}
\begin{tabularx}{\linewidth}{p{4cm} X X}
Name & BC & CP Capability \\
\midrule
Jobs range ($j$) &  $[11,12]$ & $[6,20]$ \\
Machines range ($m$) &  $[4,9]$ & $[j/2,j/1.5]$\\
Operations per job & $[3,9]$ & $[j/4,j]$\\
Options per operation & $[2,m]$ & $[1,m/1.5]$\\
Average processing time &  $[5,10]$ & $[2,4]$ \\
Deviation & 0.2 & $[1.5,3,5]$\\
\bottomrule
\end{tabularx}
\end{table}



\begin{table*}[t]
\small
\caption{Performance and average time comparison with meta-heuristic algorithms, DRL methods, OR-Tools, BC and our hybrid approach over five public datasets.}
\label{tab:allbenchmarks}
\begin{tabularx}{\textwidth}{X r r r r r r r r r r }
\multirow{2}{*}{Method} & \multicolumn{2}{c }{Brandimarte} & \multicolumn{2}{c }{Dauzere} & \multicolumn{2}{c }{edata} & \multicolumn{2}{c }{rdata} & \multicolumn{2}{c }{vdata}\\
\cmidrule(lr){2-3}
\cmidrule(lr){4-5}
\cmidrule(lr){6-7}
\cmidrule(lr){8-9}
\cmidrule(lr){10-11}
 & Gap (\%) & Time (s) & Gap (\%) & Time (s) & Gap (\%) & Time (s) & Gap (\%) & Time (s) & Gap (\%) & Time (s)\\
\midrule

IJA &  8.50 &  15.43 & 6.10 & 180.8 & 3.90 & 15.24 & 2.70 & 19.72 & 4.60 & 17.85\\
2SGA$^{*}$ &  3.17& 57.60 & – & – & – & – & – & – & 0.39 & 51.43\\
SLGA &  6.21& 283.28 & – & – & – & – & – & – & – & –\\

\hline
\cite{song2022flexible} & $27.83$ &  1.56 & $ 9.26$ &  3.00 & $ 15.53 $&  1.67 & $ 11.15$ &  1.67 & $ 4.25$ &  1.61\\
\cite{wang2023flexible}  & $ 11.97 $  & 1.23 & $ 8.88$ & 2.63 &  $ 13.73 $ & 1.37 & $ 12.17  $ & 1.35   & $ 5.40 $ & 1.34\\
\cite{echeverria2023solving} & $ 27.34 $ &  1.33 & $ 9.33 $&  2.29 & $ 17.09 $&  1.23 & $ 12.78  $&  1.26 & $ 3.38 $ &  1.62\\
\cite{ho2024residual}& $ 9.81  $ & - & -& - & $ 13.22 $& -  & $ 9.65 $ & -  & $ 3.82 $& -\\
\cite{yuan2023solving} & $ 14.04 $ &  0.41 & $ 11.10 $  &  0.80 & $ 15.54 $&  0.27 & $ 12.14  $ &  0.27 & $ 5.35  $&  0.27 \\
OR-Tools& $ 31.84  $ &  1.63 & $ 31.57 $ &  3.39 & $ 4.40$ &  1.38 & $ 11.10  $ &  1.69 & $ 28.64$ &  1.83 \\
BC&   $ 12.72$ &  1.63 & $ 7.50 $ &  4.35 & $ 12.02$ &  1.78 & $ 8.26 $ &  1.81 & $ 2.84  $ &  2.14 \\
BCxCP& $  \bm{ 8.10 }$ &  1.25 & $ \bm{ 5.15}$ &  3.22 & $ \bm{ 3.78 }$ &  0.82 & $ \bm{ 2.75 }$ &  1.32 & $ \bm{ 0.78 }$  &  1.76\\
\bottomrule
\end{tabularx}
\footnotesize{$^*$ This approach only uses the first 30 instances for the vdata benchmark.}\\
\end{table*}

As a test dataset, five sets of instances were used: Brandimarte \citep{brandimarte1993routing}, Hurink \citep{hurink1994tabu} (further divided into three groups - vdata, edata, and rdata), and Dauzère-Pérès and Paulli \citep{dauzere1994solving}. This set of benchmark instances is widely used in the literature and comprises various instance sizes, from instances with few operations to medium-large instances.

\subsubsection{Evaluation metric and baselines}

To assess the quality of solutions produced by BCxCP, the objective of minimizing the makespan has been adopted. The metric used to compare different approaches is the optimal gap. For method $m$ and instance $i$, this gap is defined as:
\begin{equation}\label{eq:gap}
g_{m_i} = \frac{(C_{max}^{m_i} - C_{max}^{o_i})}{C_{max}^{o_i}}
\end{equation}
where $C_{max}^{m_i}$ represents the makespan generated by method $m$ for instance $i$, and $C_{max}^{o_i}$ is the best or approximate optimal solution for that instance.

Our approach was compared against five recent DRL-based approaches \citep{song2022flexible, wang2023flexible, ho2024residual, yuan2023solving, echeverria2023solving}, along with OR-Tools, employing runtime constraints to ensure comparable execution times. For \cite{song2022flexible}, \cite{echeverria2023solving} and \cite{wang2023flexible}, the models available in their code repositories were used with default settings. As for \cite{ho2024residual} and \cite{yuan2023solving}, since they did not publish their code, the values reported in their papers for the benchmarks were utilized, except for the DPPaulli benchmark for \cite{ho2024residual}, which could not be generated. \cite{ho2024residual} did not provide their execution times. Furthermore, despite their lengthier execution times, we included meta-heuristic algorithms such as the Improved Jaya Algorithm (IJA) \citep{caldeira2019solving}, TwoStage Genetic Algorithm (2SGA) \citep{rooyani2019efficient}, and Self-learning Genetic Algorithm (SLGA) \citep{chen2020self}. We opted not to compare our method to dispatching rules, as the superiority of the DRL-based approaches we are evaluating has already been well-established.

\subsection{Experimental results}

In this subsection, we will present the results for the CP capability predictor, the performance of BCxCP on benchmarks, and preliminary results for the TSP.

\subsubsection{CP capability predictor results}\label{exp:cpcap}

\begin{figure*}[!t] 
	\centering 
    \includegraphics[width=\textwidth]{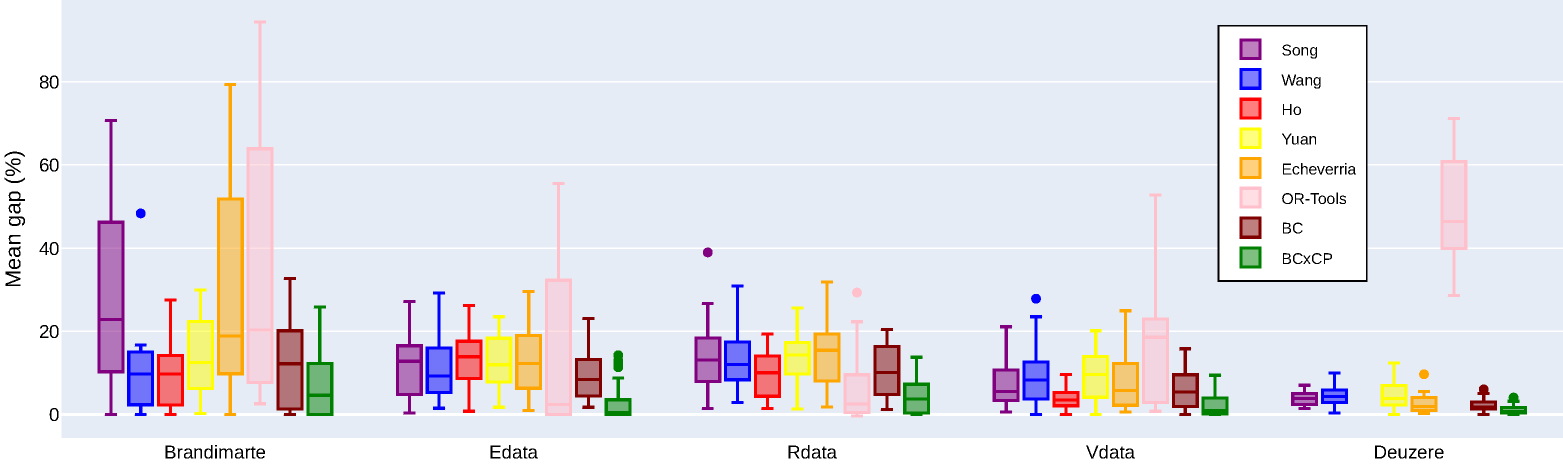} 
	\caption{Boxplots of the distribution of the mean gap for the different methods considered in this investigation on five public datasets.} 
	\label{fig:benchmarks} 
\end{figure*}

First, the results of the CP capability predictor are presented, along with an analysis of the most relevant features. As mentioned in the previous section, it is crucial that the addition of this method does not significantly slow down the inference and remains size-agnostic. Therefore, only shallow machine learning models and instance features that can be rapidly computed have been used. The chosen method was the gradient boosting regressor with default parameters.

\begin{figure}[H] 
	\centering 
    \includegraphics[width=1\linewidth]{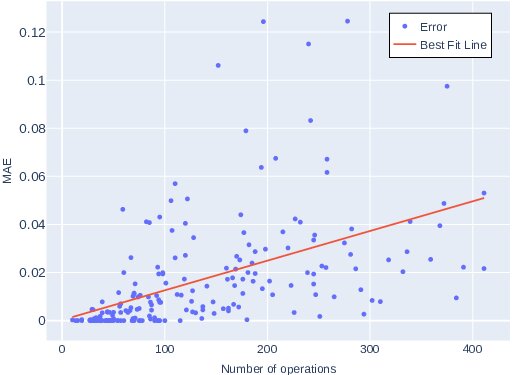} 
	\caption{MAE of the CP capability predictor as a function of the number of operations in the instance.} 
	\label{fig:errorCPcap} 
\end{figure}

The mean absolute error (MAE) obtained by our method on the validation set was 0.019. These results suggest that, with the defined predictor variables, it is possible to forecast the performance of OR-Tools for a given instance. Figure \ref{fig:errorCPcap} displays the MAE, on the y-axis, against the number of operations in the instance on the x-axis, in the validation set. It can be observed that, when the instance is small, the method is capable of predicting that OR-Tools will resolve it in real-time. The instances where the model's output is zero contribute to the low average MAE. However, the inability of our method to accurately predict on larger instances is not an issue because OR-Tools will never be utilized in such cases.

\begin{figure}[H] 
	\centering 
    \includegraphics[width=\linewidth]{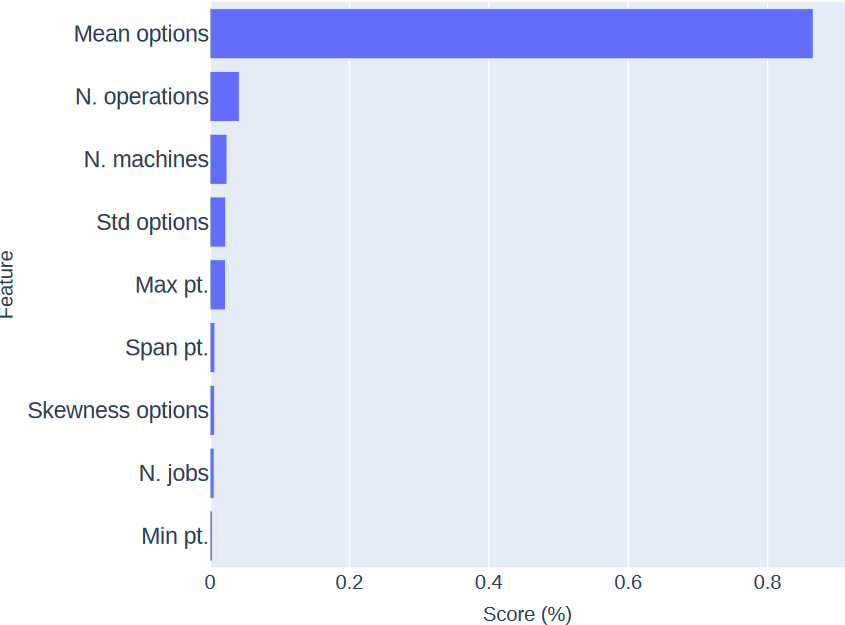} 
	\caption{Feature importance in the CP capability predictor.} 
	\label{fig:featureimportance} 
\end{figure}

We conclude this analysis by examining the importance of variables, depicted in Figure \ref{fig:featureimportance}. The importance is  determined by analyzing how much each feature reduces the  error in the model's predictions across all the trees in the ensemble. It is noticeable that the most important variable is by far the number of options an operation has, followed by the number of operations. This is sensible since these two values will influence the number of constraints in the FJSSP model as a CP and the solution space to consider.

\subsubsection{Benchmark results}

In this subsection, we empirically compare the performance of a method trained using BC and BCxCP with the baselines. In Table \ref{tab:allbenchmarks}, the mean gaps of different approaches in the benchmarks, along with their execution times, are presented. The first three rows present the results of the three meta-heuristic algorithms. The next rows show the DRL methods, OR-Tools and a model trained with BC and BCxCP. All results of these approaches across all the benchmarks are shown in \ref{appendix:vdata}. For each benchmark, the best results among the approaches that have a similar execution time is highlighted. 

Several conclusions can be drawn from the obtained results. Firstly, when comparing BC with the other five DRL-based methods the BC approach outperforms DRL-based methods in all benchmarks except for the Brandimarte benchmark with \cite{wang2023flexible} and \cite{ho2024residual}, where a slightly higher mean gap is achieved. This suggests that BC has the capability to achieve better results than DRL approaches. This could be due to the fact that, unlike DRL, which needs to explore a vast space of states and actions, BC provides the agent with only high-quality samples. Additionally, it is worth noting that only 1000 instances are used for training.

Secondly, it is observed that OR-Tools yields poor results except in the edata benchmark, mainly because the mean number of options available for operations in the edata benchmark is 1.5, indicating lower complexity. It is also evident that the worst results are obtained in the more complex instances, as can be seen in shown in \ref{appendix:vdata}. This implies that exact methods, such as OR-Tools, have a place in a dynamic FJSSP resolution methodology, complementing DL-based methods, particularly for instances with lower complexity. 

BCxCP substantially achieves better results in terms of reducing the mean gap. Out of the 148 instances analyzed, it performs the best (either being the best or tying with another method) in 113 instances. This suggest that the hybridization of methods is a promising approach for solving the FJSSP and other scheduling problems. Furthermore, there is a significant difference between BC and BCxCP, as the CP capability predictor can accurately predict when OR-Tools can be used in these instances.

Finally, it is worth noting that BCxCP is capable of achieving similar results to meta-heuristic approaches. Compared to IJA, the only method that also utilizes the same benchmarks, BCxCP outperforms this approach in all but one, where it obtains a result 0.05 worse for the rdata dataset. BCxCP yields slightly poorer results than the other two meta-heuristics, although its execution time is also much longer.

\subsection{Preliminary Results on TSP}\label{sub:tsp}

The objective of conducting experiments on another CO problem is to study if our hybrid approach can be transferable to problems beyond scheduling. In this section, we present preliminary experiments for the TSP. This problem has been chosen due to the importance of routing problems. An instance of the TSP is defined by a set $\mathcal{V} = \{v_1, v_2, ..., v_n\}$ of $n$ cities and a distance function $d : \mathcal{V} \times \mathcal{V} \rightarrow \mathbb{R}$ where $d(v_i,v_j)$ denotes the distance between cities $v_i, v_j \in \mathcal{V} $. The most common function is usually the Euclidean distance. The objective is to find the shortest possible tour that visits each city exactly once and returns to the starting city. Formally, let $\mathcal{G} = (\mathcal{V}, \mathcal{E})$ be a complete undirected graph where $V$ represents the set of cities and $\mathcal{E}$ represents the set of edges, such that for each $v_i, v_j \in \mathcal{V}$, $(v_i,v_j) \in \mathcal{E}$. The weight $w(v_i,v_j)$ of each edge $(v_i,v_j)$ is given by the distance function $d(v_i,v_j)$. A tour $\mathcal{T}$ is a Hamiltonian cycle in $\mathcal{G}$ that visits every vertex exactly once. The goal is to find a tour $T$ with minimum total weight.

\begin{figure}[H] 
	\centering 
	    \includegraphics[width=\columnwidth]{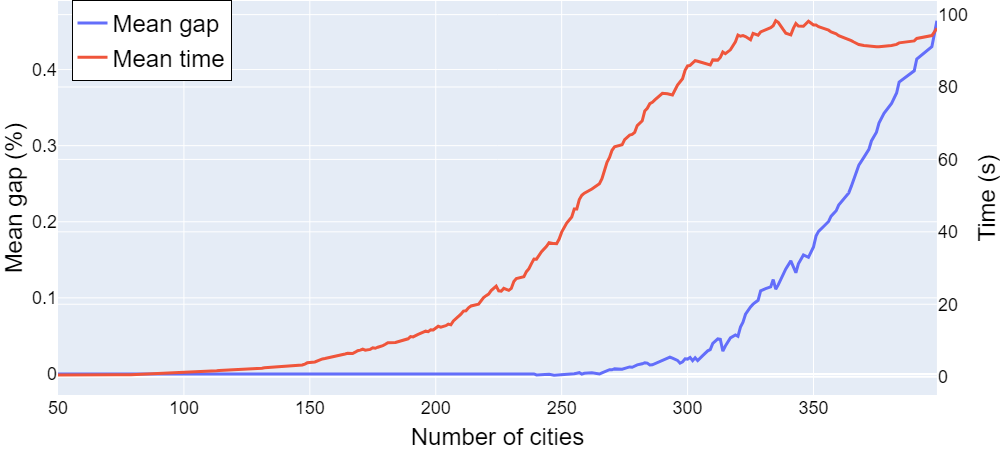} 
        \caption{Evolution of the mean gap and mean time to solve the instance as the size of the problem increases.}
        \label{fig:tsp_evolution} 
\end{figure} 

Instead of using BC to generate the policy, the well-known POMO method \citep{kwon2020pomo}, based on DRL, was employed. This paper employs three ways of generating solutions: 1) Generating a single solution by selecting the greedy action proposed by the policy. 2) Generating a set of solutions by sampling the policy. 3) This paper suggests leveraging the TSP symmetries and proposes a technique that involves not just sampling but also instance augmentation, through rotating and flipping the plane to generate another TSP instance while yielding the same result. The approach that will be compared to the hybrid method will be the latter, as it yields the best results.

The hybridization of the methods can be carried out in a similar manner to that presented for the FJSSP. Firstly, the policy is employed to begin generating a solution until a threshold is reached, after which an exact method can be used to solve the remaining instance. However, this subinstance of the problem cannot be solved like the TSP since it must start where the sub-tour ended and finish where it began, so that the final tour starts and finishes in the city of origin. Formally, if we define $\mathcal{ST}_1 = \{ v_1, v_2, \ldots , v_t \}$ as the sub-tour generated by the policy, it must be ensured that the second sub-tour starts at $v_t$ and ends at $v_1$. For this purpose, we define $\mathcal{G}^\prime = (\mathcal{V}^\prime, \mathcal{E}^\prime)$, where $\mathcal{V}^\prime = \mathcal{V} - \mathcal{ST}_1 \cup \{ v_1, v_t, v_a\}$, where $v_a$ is an auxiliary node such that $d(v_a, v_1) = d(v_a, v_t) = 0$ and $d(v_a, v_i) = \infty $ for every node $v_i \in \mathcal{V}^\prime \; | \; v_i \not\in \{v_1,v_t\}$. Solving this instance ensures that the nodes $v_1$ and $v_t$ are connected to $v_a$. Once the instance is solved, the solution will be $\mathcal{ST}_2 = \{ v_t, v_{t+1}, \ldots , v_1, v_a \}$, so by removing this node, we obtain the desired sub-tour, which can be concatenated with $\mathcal{ST}_1$ to generate a valid solution.

To model the TSP as an integer linear program, the Dantzig–Fulkerson–Johnson formulation was used and the Gurobi solver \citep{gurobi} was chosen to solve the problem. Naturally, there are much more powerful algorithms to solve the TSP, such as Concorde or LKH3 \citep{helsgaun2017extension}. Still, in this paper Gurobi is used to demonstrate the potential for improving machine learning-based methods more broadly, as it can model multiple CO problems for which the existing methods are not as effective as the ones for the TSP.

\begin{figure}[H] 
	\centering 
    \includegraphics[width=\linewidth]{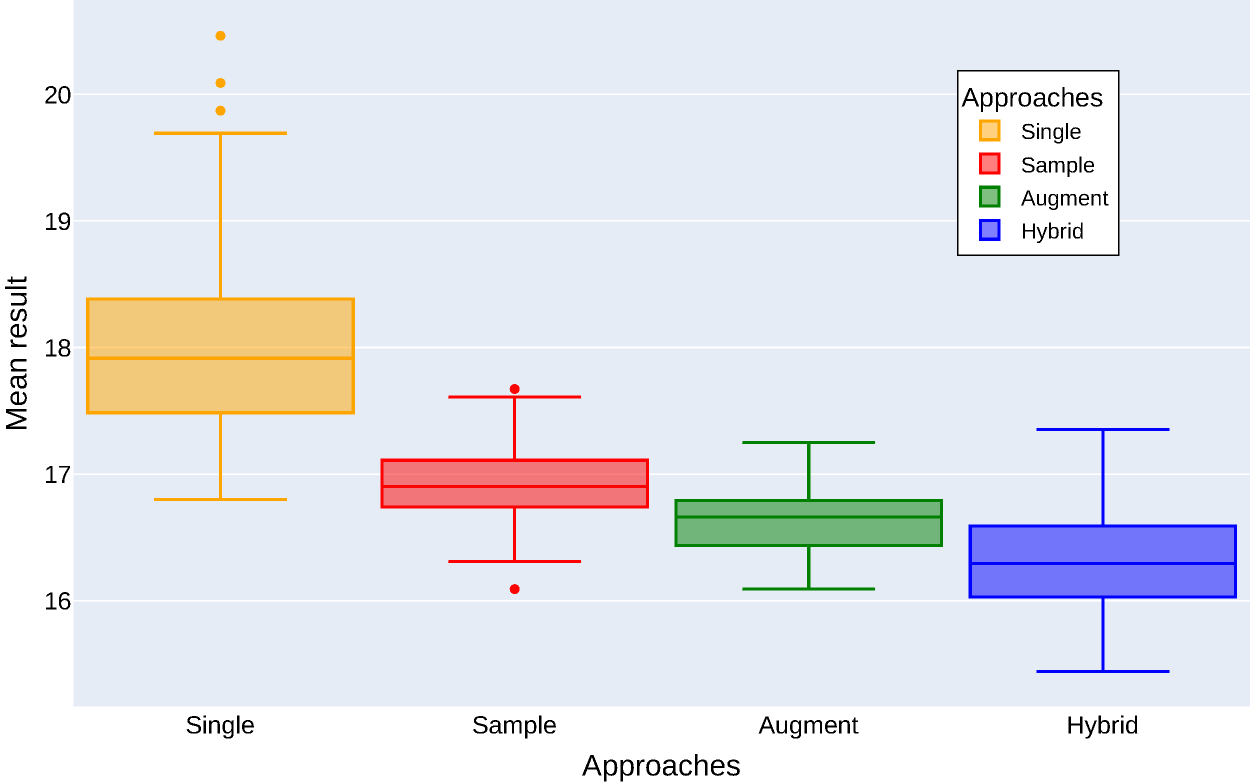}
    \caption{Results of different approaches for 100 instances of the TSP with $n=400$ cities.}
    \label{fig:tspboxplot}
\end{figure}

To define the threshold, 250 instances of the TSP were generated with a range of cities between 50 and 400, and they were solved using Gurobi with a time limit of 120 seconds. Figure \ref{fig:tsp_evolution} illustrates the relationship between the number of cities (x-axis), the gap between the upper and lower bounds found by the solver (left y-axis, shown by the blue line), and the execution time (right y-axis, indicated by the red line). The gap is defined in a similar way as in (\ref{eq:gap}). Instead of the makespan the total distance of the tour is employed.  It can be observed that beyond 100 cities, the execution time begins to increase, while the obtained gap remains stable until reaching 300 cities. The lines have been smoothed using the Savitzky–Golay filter. The threshold was chosen in such a way that the execution time was similar to that of instance augmentation, which was around 23 seconds. The threshold was set at 215 cities, with an average execution time of 20 seconds. The method for predicting when to use one method or another could be improved by taking into account features of the instance to be solved, similar to what was done with the FJSSP using the CP capability predictor.

To compare the methods, 100 instances of 400 points were randomly generated, as proposed by \cite{kool2018attention}. Figure \ref{fig:tspboxplot} displays the results of four approaches. On the x-axis, each approach is depicted: generating a single instance, performing sampling of 400 instances, augmenting the instance 8 times and performing sampling, and hybridization. The y-axis represents the mean distance obtained when solving the instances. Each instance was also solved with Gurobi with an execution time limit of 120 seconds (6 times longer than the time taken by the hybrid method and instance augmentation), but the results are not shown in the figure because they are almost 6 times worse than generating a single instance. As can be seen, the hybrid method achieves the best results, although, compared to instance augmentation, it has more variability. This is because the single trajectory method is used to build part of the solution, which has a significant variability.

\section{Conclusions and future work}\label{sec:conclusion}
This paper presents a new approach to the FJSSP by combining the strengths of BC and CP. By integrating CP with a DL framework, we utilize CP's capabilities for efficiently solving smaller instances and DL's ability to manage the complexity of larger problems. This hybrid method improves the learning process by using high-quality instances solved by CP as training data and enables a smooth transition between DL and CP to optimize the solution development for different stages of the problem.

Our evaluations across various FJSSP benchmarks show that our method outperforms five state-of-the-art DRL methods and a well-known CP solver, and it achieves competitive results compared to meta-heuristic approaches. Moreover, our initial exploration of applying this hybrid model to other combinatorial optimization problems, such as the TSP, suggests its potential for broader application.

Future work will aim to refine this hybrid model further, exploring its application to a wider range of problems, including other routing or scheduling challenges, and extending the types of methods used for algorithm hybridization, such as incorporating meta-heuristic approaches.

\section*{Acknowledgements}
This work was partially financed by the Basque Government through their Elkartek program (SONETO project, ref. KK-2023/00038) and the Gipuzkoa Provincial Council through their Gipuzkoa network of Science, Technology and Innovation program (KATEAN  project, ref. 2023-CIEN-000053-01). R. Santana thanks the Misiones Euskampus 2.0 programme for the financial help received through Euskampus Fundazioa. Partial support by the Research Groups 2022-2024 (IT1504-22) and the Elkartek Program (KK-2020/00049, KK-2022/00106, SIGZE, K-2021/00065) from the Basque Government, and the PID2019-104966GB-I00 and PID2022-137442NB-I00 research projects from the Spanish Ministry of Science is acknowledged.

\bibliographystyle{elsarticle-harv} 
\bibliography{main}

\begin{thebibliography}{51}
\expandafter\ifx\csname natexlab\endcsname\relax\def\natexlab#1{#1}\fi
\providecommand{\url}[1]{\texttt{#1}}
\providecommand{\href}[2]{#2}
\providecommand{\path}[1]{#1}
\providecommand{\DOIprefix}{doi:}
\providecommand{\ArXivprefix}{arXiv:}
\providecommand{\URLprefix}{URL: }
\providecommand{\Pubmedprefix}{pmid:}
\providecommand{\doi}[1]{\href{http://dx.doi.org/#1}{\path{#1}}}
\providecommand{\Pubmed}[1]{\href{pmid:#1}{\path{#1}}}
\providecommand{\bibinfo}[2]{#2}
\ifx\xfnm\relax \def\xfnm[#1]{\unskip,\space#1}\fi
\bibitem[{Behnke and Geiger(2012)}]{behnke2012test}
\bibinfo{author}{Behnke, D.}, \bibinfo{author}{Geiger, M.J.}, \bibinfo{year}{2012}.
\newblock \bibinfo{title}{Test instances for the flexible job shop scheduling problem with work centers}.
\newblock \bibinfo{journal}{Arbeitspapier/Research Paper/Helmut-Schmidt-Universit{\"a}t, Lehrstuhl f{\"u}r Betriebswirtschaftslehre, insbes. Logistik-Management} .
\bibitem[{Behnke and Heafield(2020)}]{behnke2020losing}
\bibinfo{author}{Behnke, M.}, \bibinfo{author}{Heafield, K.}, \bibinfo{year}{2020}.
\newblock \bibinfo{title}{Losing heads in the lottery: Pruning transformer attention in neural machine translation}, in: \bibinfo{booktitle}{Proceedings of the 2020 Conference on Empirical Methods in Natural Language Processing (EMNLP)}, pp. \bibinfo{pages}{2664--2674}.
\bibitem[{Bello et~al.(2016)Bello, Pham, Le, Norouzi and Bengio}]{bello2016neural}
\bibinfo{author}{Bello, I.}, \bibinfo{author}{Pham, H.}, \bibinfo{author}{Le, Q.V.}, \bibinfo{author}{Norouzi, M.}, \bibinfo{author}{Bengio, S.}, \bibinfo{year}{2016}.
\newblock \bibinfo{title}{Neural combinatorial optimization with reinforcement learning}.
\newblock \bibinfo{journal}{arXiv preprint arXiv:1611.09940} .
\bibitem[{Bengio et~al.(2021)Bengio, Lodi and Prouvost}]{bengio2021machine}
\bibinfo{author}{Bengio, Y.}, \bibinfo{author}{Lodi, A.}, \bibinfo{author}{Prouvost, A.}, \bibinfo{year}{2021}.
\newblock \bibinfo{title}{Machine learning for combinatorial optimization: a methodological tour d’horizon}.
\newblock \bibinfo{journal}{European Journal of Operational Research} \bibinfo{volume}{290}, \bibinfo{pages}{405--421}.
\bibitem[{Blum and Roli(2003)}]{blum2003metaheuristics}
\bibinfo{author}{Blum, C.}, \bibinfo{author}{Roli, A.}, \bibinfo{year}{2003}.
\newblock \bibinfo{title}{Metaheuristics in combinatorial optimization: Overview and conceptual comparison}.
\newblock \bibinfo{journal}{ACM Computing Surveys (CSUR)} \bibinfo{volume}{35}, \bibinfo{pages}{268--308}.
\bibitem[{Brandimarte(1993)}]{brandimarte1993routing}
\bibinfo{author}{Brandimarte, P.}, \bibinfo{year}{1993}.
\newblock \bibinfo{title}{Routing and scheduling in a flexible job shop by tabu search}.
\newblock \bibinfo{journal}{Annals of Operations Research} \bibinfo{volume}{41}, \bibinfo{pages}{157--183}.
\bibitem[{Cai et~al.(2023)Cai, Li, Luo and He}]{cai2023real}
\bibinfo{author}{Cai, L.}, \bibinfo{author}{Li, W.}, \bibinfo{author}{Luo, Y.}, \bibinfo{author}{He, L.}, \bibinfo{year}{2023}.
\newblock \bibinfo{title}{Real-time scheduling simulation optimisation of job shop in a production-logistics collaborative environment}.
\newblock \bibinfo{journal}{International Journal of Production Research} \bibinfo{volume}{61}, \bibinfo{pages}{1373--1393}.
\bibitem[{Caldeira and Gnanavelbabu(2019)}]{caldeira2019solving}
\bibinfo{author}{Caldeira, R.H.}, \bibinfo{author}{Gnanavelbabu, A.}, \bibinfo{year}{2019}.
\newblock \bibinfo{title}{Solving the flexible job shop scheduling problem using an improved jaya algorithm}.
\newblock \bibinfo{journal}{Computers \& Industrial Engineering} \bibinfo{volume}{137}, \bibinfo{pages}{106064}.
\bibitem[{Chen et~al.(2015)Chen, Wang, Chen, Gao, Xu and Nevatia}]{chen2015abc}
\bibinfo{author}{Chen, K.}, \bibinfo{author}{Wang, J.}, \bibinfo{author}{Chen, L.C.}, \bibinfo{author}{Gao, H.}, \bibinfo{author}{Xu, W.}, \bibinfo{author}{Nevatia, R.}, \bibinfo{year}{2015}.
\newblock \bibinfo{title}{{ABC-CNN}: An attention based convolutional neural network for visual question answering}.
\newblock \bibinfo{journal}{arXiv preprint arXiv:1511.05960} .
\bibitem[{Chen et~al.(2020)Chen, Yang, Li and Wang}]{chen2020self}
\bibinfo{author}{Chen, R.}, \bibinfo{author}{Yang, B.}, \bibinfo{author}{Li, S.}, \bibinfo{author}{Wang, S.}, \bibinfo{year}{2020}.
\newblock \bibinfo{title}{A self-learning genetic algorithm based on reinforcement learning for flexible job-shop scheduling problem}.
\newblock \bibinfo{journal}{Computers \& Industrial Engineering} \bibinfo{volume}{149}, \bibinfo{pages}{106778}.
\bibitem[{Chen et~al.(2023)Chen, Khir and Van~Hentenryck}]{chen2023two}
\bibinfo{author}{Chen, W.}, \bibinfo{author}{Khir, R.}, \bibinfo{author}{Van~Hentenryck, P.}, \bibinfo{year}{2023}.
\newblock \bibinfo{title}{Two-stage learning for the flexible job shop scheduling problem}.
\newblock \bibinfo{journal}{arXiv preprint arXiv:2301.09703} .
\bibitem[{Czarnecki et~al.(2019)Czarnecki, Pascanu, Osindero, Jayakumar, Swirszcz and Jaderberg}]{czarnecki2019distilling}
\bibinfo{author}{Czarnecki, W.M.}, \bibinfo{author}{Pascanu, R.}, \bibinfo{author}{Osindero, S.}, \bibinfo{author}{Jayakumar, S.}, \bibinfo{author}{Swirszcz, G.}, \bibinfo{author}{Jaderberg, M.}, \bibinfo{year}{2019}.
\newblock \bibinfo{title}{Distilling policy distillation}, in: \bibinfo{booktitle}{The 22nd International Conference on Artificial Intelligence and Statistics}, \bibinfo{organization}{PMLR}. pp. \bibinfo{pages}{1331--1340}.
\bibitem[{Da~Col and Teppan(2019a)}]{da2019google}
\bibinfo{author}{Da~Col, G.}, \bibinfo{author}{Teppan, E.}, \bibinfo{year}{2019}a.
\newblock \bibinfo{title}{{Google vs IBM}: A constraint solving challenge on the job-shop scheduling problem}.
\newblock \bibinfo{journal}{arXiv preprint arXiv:1909.08247} .
\bibitem[{Da~Col and Teppan(2019b)}]{da2019industrial}
\bibinfo{author}{Da~Col, G.}, \bibinfo{author}{Teppan, E.C.}, \bibinfo{year}{2019}b.
\newblock \bibinfo{title}{Industrial size job shop scheduling tackled by present day {CP} solvers}, in: \bibinfo{booktitle}{Principles and Practice of Constraint Programming: 25th International Conference, CP 2019, Stamford, CT, USA, September 30--October 4, 2019, Proceedings 25}, \bibinfo{organization}{Springer}. pp. \bibinfo{pages}{144--160}.
\bibitem[{Dauz{\`e}re-P{\'e}r{\`e}s and Paulli(1994)}]{dauzere1994solving}
\bibinfo{author}{Dauz{\`e}re-P{\'e}r{\`e}s, S.}, \bibinfo{author}{Paulli, J.}, \bibinfo{year}{1994}.
\newblock \bibinfo{title}{Solving the general multiprocessor job-shop scheduling problem}.
\newblock \bibinfo{journal}{Management Report Series} \bibinfo{volume}{182}.
\bibitem[{Echeverria et~al.(2023)Echeverria, Murua and Santana}]{echeverria2023solving}
\bibinfo{author}{Echeverria, I.}, \bibinfo{author}{Murua, M.}, \bibinfo{author}{Santana, R.}, \bibinfo{year}{2023}.
\newblock \bibinfo{title}{Solving large flexible job shop scheduling instances by generating a diverse set of scheduling policies with deep reinforcement learning}.
\newblock \bibinfo{journal}{arXiv preprint arXiv:2310.15706} .
\bibitem[{Fey and Lenssen(2019)}]{Fey/Lenssen/2019}
\bibinfo{author}{Fey, M.}, \bibinfo{author}{Lenssen, J.E.}, \bibinfo{year}{2019}.
\newblock \bibinfo{title}{Fast graph representation learning with {PyTorch Geometric}}, in: \bibinfo{booktitle}{ICLR Workshop on Representation Learning on Graphs and Manifolds}.
\bibitem[{Garey et~al.(1976)Garey, Johnson and Sethi}]{garey1976complexity}
\bibinfo{author}{Garey, M.R.}, \bibinfo{author}{Johnson, D.S.}, \bibinfo{author}{Sethi, R.}, \bibinfo{year}{1976}.
\newblock \bibinfo{title}{The complexity of flowshop and jobshop scheduling}.
\newblock \bibinfo{journal}{Mathematics of Operations Research} \bibinfo{volume}{1}, \bibinfo{pages}{117--129}.
\bibitem[{Garrido et~al.(2000)Garrido, Salido, Barber and L{\'o}pez}]{garrido2000heuristic}
\bibinfo{author}{Garrido, A.}, \bibinfo{author}{Salido, M.A.}, \bibinfo{author}{Barber, F.}, \bibinfo{author}{L{\'o}pez, M.}, \bibinfo{year}{2000}.
\newblock \bibinfo{title}{Heuristic methods for solving job-shop scheduling problems}, in: \bibinfo{booktitle}{Proc. ECAI-2000 Workshop on New Results in Planning, Scheduling and Design (PuK2000)}, pp. \bibinfo{pages}{44--49}.
\bibitem[{{Gurobi Optimization, LLC}(2023)}]{gurobi}
\bibinfo{author}{{Gurobi Optimization, LLC}}, \bibinfo{year}{2023}.
\newblock \bibinfo{title}{{Gurobi Optimizer Reference Manual}}.
\newblock \URLprefix \url{https://www.gurobi.com}.
\bibitem[{Helsgaun(2017)}]{helsgaun2017extension}
\bibinfo{author}{Helsgaun, K.}, \bibinfo{year}{2017}.
\newblock \bibinfo{title}{{An extension of the Lin-Kernighan-Helsgaun TSP solver for constrained traveling salesman and vehicle routing problems}}.
\newblock \bibinfo{journal}{Roskilde: Roskilde University} \bibinfo{volume}{12}.
\bibitem[{Ho et~al.(2024)Ho, Cheng, Wu, Chiang, Chen, Wu and Wu}]{ho2024residual}
\bibinfo{author}{Ho, K.H.}, \bibinfo{author}{Cheng, J.Y.}, \bibinfo{author}{Wu, J.H.}, \bibinfo{author}{Chiang, F.}, \bibinfo{author}{Chen, Y.C.}, \bibinfo{author}{Wu, Y.Y.}, \bibinfo{author}{Wu, I.C.}, \bibinfo{year}{2024}.
\newblock \bibinfo{title}{Residual scheduling: A new reinforcement learning approach to solving job shop scheduling problem}.
\newblock \bibinfo{journal}{IEEE Access} .
\bibitem[{Homayouni and Fontes(2021)}]{homayouni2021production}
\bibinfo{author}{Homayouni, S.M.}, \bibinfo{author}{Fontes, D.B.}, \bibinfo{year}{2021}.
\newblock \bibinfo{title}{Production and transport scheduling in flexible job shop manufacturing systems}.
\newblock \bibinfo{journal}{Journal of Global Optimization} \bibinfo{volume}{79}, \bibinfo{pages}{463--502}.
\bibitem[{Hu et~al.(2020)Hu, Dong, Wang and Sun}]{hu2020heterogeneous}
\bibinfo{author}{Hu, Z.}, \bibinfo{author}{Dong, Y.}, \bibinfo{author}{Wang, K.}, \bibinfo{author}{Sun, Y.}, \bibinfo{year}{2020}.
\newblock \bibinfo{title}{Heterogeneous graph transformer}, in: \bibinfo{booktitle}{Proceedings of the Web Conference 2020}, pp. \bibinfo{pages}{2704--2710}.
\bibitem[{Hurink et~al.(1994)Hurink, Jurisch and Thole}]{hurink1994tabu}
\bibinfo{author}{Hurink, J.}, \bibinfo{author}{Jurisch, B.}, \bibinfo{author}{Thole, M.}, \bibinfo{year}{1994}.
\newblock \bibinfo{title}{Tabu search for the job-shop scheduling problem with multi-purpose machines}.
\newblock \bibinfo{journal}{Operations-Research-Spektrum} \bibinfo{volume}{15}, \bibinfo{pages}{205--215}.
\bibitem[{Ingimundardottir and Runarsson(2018)}]{ingimundardottir2018discovering}
\bibinfo{author}{Ingimundardottir, H.}, \bibinfo{author}{Runarsson, T.P.}, \bibinfo{year}{2018}.
\newblock \bibinfo{title}{Discovering dispatching rules from data using imitation learning: A case study for the job-shop problem}.
\newblock \bibinfo{journal}{Journal of Scheduling} \bibinfo{volume}{21}, \bibinfo{pages}{413--428}.
\bibitem[{Kool et~al.(2018)Kool, Van~Hoof and Welling}]{kool2018attention}
\bibinfo{author}{Kool, W.}, \bibinfo{author}{Van~Hoof, H.}, \bibinfo{author}{Welling, M.}, \bibinfo{year}{2018}.
\newblock \bibinfo{title}{Attention, learn to solve routing problems!}
\newblock \bibinfo{journal}{arXiv preprint arXiv:1803.08475} .
\bibitem[{Kwon et~al.(2020)Kwon, Choo, Kim, Yoon, Gwon and Min}]{kwon2020pomo}
\bibinfo{author}{Kwon, Y.D.}, \bibinfo{author}{Choo, J.}, \bibinfo{author}{Kim, B.}, \bibinfo{author}{Yoon, I.}, \bibinfo{author}{Gwon, Y.}, \bibinfo{author}{Min, S.}, \bibinfo{year}{2020}.
\newblock \bibinfo{title}{{POMO}: Policy optimization with multiple optima for reinforcement learning}.
\newblock \bibinfo{journal}{Advances in Neural Information Processing Systems} \bibinfo{volume}{33}, \bibinfo{pages}{21188--21198}.
\bibitem[{Laborie(2009)}]{laborie2009ibm}
\bibinfo{author}{Laborie, P.}, \bibinfo{year}{2009}.
\newblock \bibinfo{title}{{IBM ILOG CP} optimizer for detailed scheduling illustrated on three problems}, in: \bibinfo{booktitle}{Integration of AI and OR Techniques in Constraint Programming for Combinatorial Optimization Problems: 6th International Conference, CPAIOR 2009 Pittsburgh, PA, USA, May 27-31, 2009 Proceedings 6}, \bibinfo{organization}{Springer}. pp. \bibinfo{pages}{148--162}.
\bibitem[{Lei et~al.(2022)Lei, Guo, Zhao, Wang, Qian, Meng and Tang}]{lei2022multi}
\bibinfo{author}{Lei, K.}, \bibinfo{author}{Guo, P.}, \bibinfo{author}{Zhao, W.}, \bibinfo{author}{Wang, Y.}, \bibinfo{author}{Qian, L.}, \bibinfo{author}{Meng, X.}, \bibinfo{author}{Tang, L.}, \bibinfo{year}{2022}.
\newblock \bibinfo{title}{A multi-action deep reinforcement learning framework for flexible job-shop scheduling problem}.
\newblock \bibinfo{journal}{Expert Systems with Applications} \bibinfo{volume}{205}, \bibinfo{pages}{117796}.
\bibitem[{Li et~al.(2022)Li, Liang, Zhu, Cao, Ding, Zha and Wu}]{li2022learning}
\bibinfo{author}{Li, L.}, \bibinfo{author}{Liang, S.}, \bibinfo{author}{Zhu, Z.}, \bibinfo{author}{Cao, X.}, \bibinfo{author}{Ding, C.}, \bibinfo{author}{Zha, H.}, \bibinfo{author}{Wu, B.}, \bibinfo{year}{2022}.
\newblock \bibinfo{title}{Learning to optimize permutation flow shop scheduling via graph-based imitation learning}.
\newblock \bibinfo{journal}{arXiv preprint arXiv:2210.17178} .
\bibitem[{Lodi and Zarpellon(2017)}]{lodi2017learning}
\bibinfo{author}{Lodi, A.}, \bibinfo{author}{Zarpellon, G.}, \bibinfo{year}{2017}.
\newblock \bibinfo{title}{On learning and branching: a survey}.
\newblock \bibinfo{journal}{TOP} \bibinfo{volume}{25}, \bibinfo{pages}{207--236}.
\bibitem[{Mazyavkina et~al.(2021)Mazyavkina, Sviridov, Ivanov and Burnaev}]{mazyavkina2021reinforcement}
\bibinfo{author}{Mazyavkina, N.}, \bibinfo{author}{Sviridov, S.}, \bibinfo{author}{Ivanov, S.}, \bibinfo{author}{Burnaev, E.}, \bibinfo{year}{2021}.
\newblock \bibinfo{title}{Reinforcement learning for combinatorial optimization: A survey}.
\newblock \bibinfo{journal}{Computers \& Operations Research} \bibinfo{volume}{134}, \bibinfo{pages}{105400}.
\bibitem[{M{\"u}ller et~al.(2022)M{\"u}ller, M{\"u}ller, Kress and Pesch}]{muller2022algorithm}
\bibinfo{author}{M{\"u}ller, D.}, \bibinfo{author}{M{\"u}ller, M.G.}, \bibinfo{author}{Kress, D.}, \bibinfo{author}{Pesch, E.}, \bibinfo{year}{2022}.
\newblock \bibinfo{title}{An algorithm selection approach for the flexible job shop scheduling problem: Choosing constraint programming solvers through machine learning}.
\newblock \bibinfo{journal}{European Journal of Operational Research} \bibinfo{volume}{302}, \bibinfo{pages}{874--891}.
\bibitem[{Nazari et~al.(2018)Nazari, Oroojlooy, Snyder and Tak{\'a}c}]{nazari2018reinforcement}
\bibinfo{author}{Nazari, M.}, \bibinfo{author}{Oroojlooy, A.}, \bibinfo{author}{Snyder, L.}, \bibinfo{author}{Tak{\'a}c, M.}, \bibinfo{year}{2018}.
\newblock \bibinfo{title}{Reinforcement learning for solving the vehicle routing problem}.
\newblock \bibinfo{journal}{Advances in Neural Information Processing Systems} \bibinfo{volume}{31}.
\bibitem[{Paschos(2014)}]{paschos2014applications}
\bibinfo{author}{Paschos, V.T.}, \bibinfo{year}{2014}.
\newblock \bibinfo{title}{Applications of combinatorial optimization}. volume~\bibinfo{volume}{3}.
\newblock \bibinfo{publisher}{John Wiley \& Sons}.
\bibitem[{Pedregosa et~al.(2011)Pedregosa, Varoquaux, Gramfort, Michel, Thirion, Grisel, Blondel, Prettenhofer, Weiss, Dubourg, Vanderplas, Passos, Cournapeau, Brucher, Perrot and Duchesnay}]{scikit-learn}
\bibinfo{author}{Pedregosa, F.}, \bibinfo{author}{Varoquaux, G.}, \bibinfo{author}{Gramfort, A.}, \bibinfo{author}{Michel, V.}, \bibinfo{author}{Thirion, B.}, \bibinfo{author}{Grisel, O.}, \bibinfo{author}{Blondel, M.}, \bibinfo{author}{Prettenhofer, P.}, \bibinfo{author}{Weiss, R.}, \bibinfo{author}{Dubourg, V.}, \bibinfo{author}{Vanderplas, J.}, \bibinfo{author}{Passos, A.}, \bibinfo{author}{Cournapeau, D.}, \bibinfo{author}{Brucher, M.}, \bibinfo{author}{Perrot, M.}, \bibinfo{author}{Duchesnay, E.}, \bibinfo{year}{2011}.
\newblock \bibinfo{title}{Scikit-learn: Machine learning in {P}ython}.
\newblock \bibinfo{journal}{Journal of Machine Learning Research} \bibinfo{volume}{12}, \bibinfo{pages}{2825--2830}.
\bibitem[{Prud'homme and Fages(2022)}]{Prud'homme2022}
\bibinfo{author}{Prud'homme, C.}, \bibinfo{author}{Fages, J.G.}, \bibinfo{year}{2022}.
\newblock \bibinfo{title}{Choco-solver: A java library for constraint programming}.
\newblock \bibinfo{journal}{Journal of Open Source Software} \bibinfo{volume}{7}, \bibinfo{pages}{4708}.
\newblock \URLprefix \url{https://doi.org/10.21105/joss.04708}, \DOIprefix\doi{10.21105/joss.04708}.
\bibitem[{Rooyani and Defersha(2019)}]{rooyani2019efficient}
\bibinfo{author}{Rooyani, D.}, \bibinfo{author}{Defersha, F.M.}, \bibinfo{year}{2019}.
\newblock \bibinfo{title}{An efficient two-stage genetic algorithm for flexible job-shop scheduling}.
\newblock \bibinfo{journal}{IFAC-PapersOnLine} \bibinfo{volume}{52}, \bibinfo{pages}{2519--2524}.
\bibitem[{Sharma and Sharma(2021)}]{sharma2021improved}
\bibinfo{author}{Sharma, M.}, \bibinfo{author}{Sharma, S.}, \bibinfo{year}{2021}.
\newblock \bibinfo{title}{An improved {NEH} heuristic to minimize makespan for flow shop scheduling problems}.
\newblock \bibinfo{journal}{Decision Science Letters} \bibinfo{volume}{10}, \bibinfo{pages}{311--322}.
\bibitem[{Shi et~al.(2020)Shi, Huang, Feng, Zhong, Wang and Sun}]{shi2020masked}
\bibinfo{author}{Shi, Y.}, \bibinfo{author}{Huang, Z.}, \bibinfo{author}{Feng, S.}, \bibinfo{author}{Zhong, H.}, \bibinfo{author}{Wang, W.}, \bibinfo{author}{Sun, Y.}, \bibinfo{year}{2020}.
\newblock \bibinfo{title}{Masked label prediction: Unified message passing model for semi-supervised classification}.
\newblock \bibinfo{journal}{arXiv preprint arXiv:2009.03509} .
\bibitem[{Song et~al.(2022)Song, Chen, Li and Cao}]{song2022flexible}
\bibinfo{author}{Song, W.}, \bibinfo{author}{Chen, X.}, \bibinfo{author}{Li, Q.}, \bibinfo{author}{Cao, Z.}, \bibinfo{year}{2022}.
\newblock \bibinfo{title}{Flexible job-shop scheduling via graph neural network and deep reinforcement learning}.
\newblock \bibinfo{journal}{IEEE Transactions on Industrial Informatics} \bibinfo{volume}{19}, \bibinfo{pages}{1600--1610}.
\bibitem[{Tassel et~al.(2023)Tassel, Gebser and Schekotihin}]{tassel2023end}
\bibinfo{author}{Tassel, P.}, \bibinfo{author}{Gebser, M.}, \bibinfo{author}{Schekotihin, K.}, \bibinfo{year}{2023}.
\newblock \bibinfo{title}{An end-to-end reinforcement learning approach for job-shop scheduling problems based on constraint programming}.
\newblock \bibinfo{journal}{arXiv preprint arXiv:2306.05747} .
\bibitem[{Vaswani et~al.(2017)Vaswani, Shazeer, Parmar, Uszkoreit, Jones, Gomez, Kaiser and Polosukhin}]{vaswani2017attention}
\bibinfo{author}{Vaswani, A.}, \bibinfo{author}{Shazeer, N.}, \bibinfo{author}{Parmar, N.}, \bibinfo{author}{Uszkoreit, J.}, \bibinfo{author}{Jones, L.}, \bibinfo{author}{Gomez, A.N.}, \bibinfo{author}{Kaiser, {\L}.}, \bibinfo{author}{Polosukhin, I.}, \bibinfo{year}{2017}.
\newblock \bibinfo{title}{Attention is all you need}.
\newblock \bibinfo{journal}{Advances in {Neural} information processing systems} \bibinfo{volume}{30}.
\bibitem[{Vinyals et~al.(2015)Vinyals, Fortunato and Jaitly}]{vinyals2015pointer}
\bibinfo{author}{Vinyals, O.}, \bibinfo{author}{Fortunato, M.}, \bibinfo{author}{Jaitly, N.}, \bibinfo{year}{2015}.
\newblock \bibinfo{title}{Pointer networks}.
\newblock \bibinfo{journal}{Advances in Neural Information Processing Systems} \bibinfo{volume}{28}.
\bibitem[{Wang et~al.(2023)Wang, Wang, Sun, Deng and Chen}]{wang2023flexible}
\bibinfo{author}{Wang, R.}, \bibinfo{author}{Wang, G.}, \bibinfo{author}{Sun, J.}, \bibinfo{author}{Deng, F.}, \bibinfo{author}{Chen, J.}, \bibinfo{year}{2023}.
\newblock \bibinfo{title}{Flexible job shop scheduling via dual attention network based reinforcement learning}.
\newblock \bibinfo{journal}{arXiv preprint arXiv:2305.05119} .
\bibitem[{Williams(1992)}]{williams1992simple}
\bibinfo{author}{Williams, R.J.}, \bibinfo{year}{1992}.
\newblock \bibinfo{title}{Simple statistical gradient-following algorithms for connectionist reinforcement learning}.
\newblock \bibinfo{journal}{Machine Learning} \bibinfo{volume}{8}, \bibinfo{pages}{229--256}.
\bibitem[{Xie et~al.(2019)Xie, Gao, Peng, Li and Li}]{xie2019review}
\bibinfo{author}{Xie, J.}, \bibinfo{author}{Gao, L.}, \bibinfo{author}{Peng, K.}, \bibinfo{author}{Li, X.}, \bibinfo{author}{Li, H.}, \bibinfo{year}{2019}.
\newblock \bibinfo{title}{Review on flexible job shop scheduling}.
\newblock \bibinfo{journal}{IET Collaborative Intelligent Manufacturing} \bibinfo{volume}{1}, \bibinfo{pages}{67--77}.
\bibitem[{Yuan et~al.(2023)Yuan, Wang, Cheng, Song, Fan and Li}]{yuan2023solving}
\bibinfo{author}{Yuan, E.}, \bibinfo{author}{Wang, L.}, \bibinfo{author}{Cheng, S.}, \bibinfo{author}{Song, S.}, \bibinfo{author}{Fan, W.}, \bibinfo{author}{Li, Y.}, \bibinfo{year}{2023}.
\newblock \bibinfo{title}{Solving flexible job shop scheduling problems via deep reinforcement learning}.
\newblock \bibinfo{journal}{Expert Systems with Applications} , \bibinfo{pages}{123019}.
\bibitem[{Zhang et~al.(2020)Zhang, Song, Cao, Zhang, Tan and Chi}]{zhang2020learning}
\bibinfo{author}{Zhang, C.}, \bibinfo{author}{Song, W.}, \bibinfo{author}{Cao, Z.}, \bibinfo{author}{Zhang, J.}, \bibinfo{author}{Tan, P.S.}, \bibinfo{author}{Chi, X.}, \bibinfo{year}{2020}.
\newblock \bibinfo{title}{Learning to dispatch for job shop scheduling via deep reinforcement learning}.
\newblock \bibinfo{journal}{Advances in Neural Information Processing Systems} \bibinfo{volume}{33}, \bibinfo{pages}{1621--1632}.
\bibitem[{Zhang et~al.(2023)Zhang, He, Chan and Chow}]{zhang2023deepmag}
\bibinfo{author}{Zhang, J.D.}, \bibinfo{author}{He, Z.}, \bibinfo{author}{Chan, W.H.}, \bibinfo{author}{Chow, C.Y.}, \bibinfo{year}{2023}.
\newblock \bibinfo{title}{{DeepMAG}: Deep reinforcement learning with multi-agent graphs for flexible job shop scheduling}.
\newblock \bibinfo{journal}{Knowledge-Based Systems} \bibinfo{volume}{259}, \bibinfo{pages}{110083}.

\end{thebibliography}

\appendix
\section{Node and edge features}\label{appendix:nodefatures}
In this appendix, the features of the nodes and edges of the states are described. For operation-type nodes, these features include:

\begin{itemize}
\item A binary value (0 or 1) indicating whether the operation can be performed.
\item The sum of the average processing times of the remaining operations. If an operation has more than one option, the average is calculated.
\item The mean processing time of that operation.
\item The minimum processing time of that operation.
\item The mean processing time divided by the average processing times of the remaining operations.
\end{itemize}

For machines, the features are:

\begin{itemize}
\item The instant at which the last assigned operation on the machine is completed.
\item The utilization percentage of that machine, defined as the ratio between the time it has been processing operations and the previous feature.
\item The difference between the time at which the machine finishes its last operation and the minimum among all machines, indicating if a machine is significantly ahead of the rest.
\end{itemize}

For jobs, the features are:

\begin{itemize}
\item A binary value (0 or 1) indicating whether the job has been completed.
\item The instant at which the last assigned operation from that job has finished.
\item The number of operations left to be performed.
\item The sum of the average processing times of the remaining operations for that job.
\end{itemize}

Regarding the edges, only two types have features: edges between operations and machines and between jobs and machines. The features for the former are:

\begin{itemize}
\item The processing time of that operation on that machine.
\item The processing time divided by the options that operation has.
\item The processing time divided by the options that machine has.
\item The processing time divided by the sum of the average processing times.
\end{itemize}

Concerning the features between machines and jobs, they are analogous to the former, except that instead of the processing time, the gap generated by assigning that operation to that machine is considered. When attempting to assign an operation to that machine, it may not be able to be executed immediately since it is necessary to wait for its preceding operation to finish, and this instant may be later than when the machine is available, creating a gap in which the machine is idle.

\end{multicols*}

\onecolumn
\section{Benchmark results}\label{appendix:vdata}
In this appendix, the results for all methods across all instances of the public benchmarks are presented. The upper bounds (UB) are obtained from \cite{behnke2012test}.
\begin{tiny}\setlength\tabcolsep{4pt}
\begin{table}[H]
\centering
\caption{Comparison of the best makespan obtained by the methods proposed by \cite{song2022flexible},  \cite{wang2023flexible}, \cite{ho2024residual}, \cite{yuan2023solving}, \cite{echeverria2023solving}, OR-Tools, BC and BCxCP in the vdata dataset.} \label{tab:vdataall} 
\begin{tabular}{l|r|r|r|r|r|r|r|r|r}
Name &           Song &           Wang &             Ho &     Echeverria &  Yuan &            OR &             BC &           BCxCP &  UB \\
\hline
\hline
Vdata04 &          579 &          620 &          620 &           599 &          612 &          577 &          589 &  \textbf{574} &  570 \\
Vdata05 &          579 &          572 &          564 &           559 &          601 & \textbf{533} &          572 &  \textbf{533} &  529 \\
Vdata06 &          513 &          525 &          492 &           525 &          533 &          484 &          502 &  \textbf{483} &  477 \\
Vdata07 &          529 &          545 &          529 &           562 &          556 &          508 &          548 &  \textbf{503} &  502 \\
Vdata08 &          507 &          490 &          481 &           470 &          525 &          464 &          491 &  \textbf{462} &  457 \\
Vdata09 &          823 &          831 &          855 &           832 &          888 &          830 &          823 &  \textbf{802} &  799 \\
Vdata10 &          785 &          828 &          757 &           792 &          802 &          765 &          774 &  \textbf{752} &  749 \\
Vdata11 &          806 &          810 &          778 &           779 &          783 &          778 &          780 &  \textbf{765} &  765 \\
Vdata12 &          871 &          886 &          869 &           867 &          887 &          873 &          873 &  \textbf{855} &  853 \\
Vdata13 &          843 &          831 &          840 &           818 &          843 &          815 &          848 &  \textbf{805} &  804 \\
Vdata14 &         1122 &         1120 &         1086 &          1103 &         1119 &         1141 &         1082 & \textbf{1072} & 1071 \\
Vdata15 &          950 &          959 &          974 &           951 &          977 &          967 &          949 &  \textbf{936} &  936 \\
Vdata16 &         1065 &         1063 &         1074 &          1050 &         1082 &         1074 &         1056 & \textbf{1038} & 1038 \\
Vdata17 &         1108 &         1099 &         1098 &          1087 &         1088 &         1145 &         1079 & \textbf{1071} & 1070 \\
Vdata18 &         1144 &         1106 &         1112 &          1095 &         1110 &         1100 &         1094 & \textbf{1090} & 1089 \\
Vdata19 &          764 &          752 &          740 &           735 &          727 &          874 &          728 &  \textbf{717} &  717 \\
Vdata20 &          659 &          728 &          708 &           650 &          648 &          781 & \textbf{646} &  \textbf{646} &  646 \\
Vdata21 &          667 & \textbf{663} & \textbf{663} &           674 & \textbf{663} &          787 &          670 &           670 &  663 \\
Vdata22 &          660 &          641 &          671 &           694 &          639 &          896 &          653 &  \textbf{624} &  617 \\
Vdata23 &          779 &          784 & \textbf{756} &           770 &          766 &          911 & \textbf{756} &  \textbf{756} &  756 \\
Vdata24 & \textbf{830} &          947 &          847 &           834 &          891 &         1231 &          854 &           832 &  806 \\
Vdata25 &          787 &          816 &          765 &           782 &          809 &          994 &          763 &  \textbf{746} &  739 \\
Vdata26 &          849 &          876 &          858 &           848 &          872 &         1336 &          864 &  \textbf{831} &  815 \\
Vdata27 &          816 &          826 &          828 &           817 &          873 &         1330 &          800 &  \textbf{793} &  777 \\
Vdata28 &          809 &          831 &          827 &           829 &          840 &         1254 &          794 &  \textbf{780} &  756 \\
Vdata29 &         1079 &         1105 &         1080 &          1063 &         1100 &         1497 &         1070 & \textbf{1057} & 1054 \\
Vdata30 &         1136 &         1137 &         1110 &          1112 &         1131 &         1623 &         1109 & \textbf{1092} & 1085 \\
Vdata31 &         1124 &         1101 &         1095 &          1098 &         1106 &         1445 &         1088 & \textbf{1074} & 1070 \\
Vdata32 &         1029 &         1038 &         1031 &          1007 &         1076 &         1410 &         1036 & \textbf{1002} &  994 \\
Vdata33 &         1111 &         1113 &         1119 &          1117 &         1143 &         1668 &         1092 & \textbf{1074} & 1069 \\
Vdata34 &         1547 &         1586 &         1561 &          1527 &         1575 &         1955 &         1540 & \textbf{1521} & 1520 \\
Vdata35 &         1706 &         1687 &         1702 &          1676 &         1726 &         2317 &         1700 & \textbf{1664} & 1658 \\
Vdata36 &         1533 &         1539 &         1527 &          1511 &         1531 &         2107 &         1503 & \textbf{1500} & 1497 \\
Vdata37 &         1579 &         1583 &         1548 &          1553 &         1594 &         2104 &         1553 & \textbf{1538} & 1535 \\
Vdata38 &         1606 &         1615 &         1603 &          1570 &         1575 &         2168 &         1566 & \textbf{1557} & 1549 \\
Vdata39 &          961 &          987 &          982 &  \textbf{950} &          973 &         1519 &          964 &           964 &  948 \\
Vdata40 &         1041 &         1055 &         1044 & \textbf{1018} &         1024 &         1585 &         1026 &          1026 &  986 \\
Vdata41 &          996 &          946 & \textbf{943} &           946 & \textbf{943} &         1350 & \textbf{943} &  \textbf{943} &  943 \\
Vdata42 &          936 &          976 &          964 &           973 &          942 &         1508 &          939 &  \textbf{932} &  922 \\
Vdata43 &         1008 & \textbf{961} &          966 &           986 &          967 &         1461 &          968 &           968 &  955 \\
\hline
Gap (\%) & $ 4.25$ & $ 5.4$ & $ 3.18$ & $ 3.38 $ & $ 5.35$ & $ 28.64$ & $ 2.84 $ & $ \bm{ 0.78}$
\end{tabular}
\end{table}
\end{tiny}

\begin{tiny}\setlength\tabcolsep{4pt}
\begin{table}[H]
\small
\centering
\caption{Comparison of the best makespan obtained by the methods proposed by \cite{song2022flexible},  \cite{wang2023flexible}, \cite{ho2024residual}, \cite{yuan2023solving}, \cite{echeverria2023solving}, OR-Tools, BC and BCxCP in the edata dataset.} \label{tab:edataeall} 
\begin{tabular}{l|r|r|r|r|r|r|r|r|r}
Name &           Song &           Wang &             Ho &     Echeverria &  Yuan &            OR &             BC &           BCxCP &  UB \\
\hline
\hline
Edata04 &  688 &           665 &  658 &       725 &  658 &  \textbf{609} &  666 &  \textbf{609} &  609 \\
Edata05 &  761 &           761 &  789 &       849 &  809 &  \textbf{655} &  806 &  \textbf{655} &  655 \\
Edata06 &  657 &           632 &  611 &       663 &  660 &  \textbf{550} &  611 &  \textbf{550} &  550 \\
Edata07 &  678 &           734 &  679 &       703 &  669 &  \textbf{568} &  663 &  \textbf{568} &  568 \\
Edata08 &  530 &           530 &  530 &       602 &  593 &  \textbf{503} &  538 &  \textbf{503} &  503 \\
Edata09 &  952 &           899 &  875 &       875 &  951 &  \textbf{833} &  875 &  \textbf{833} &  833 \\
Edata10 &  969 &           840 &  868 &       834 &  896 &  \textbf{762} &  853 &  \textbf{762} &  762 \\
Edata11 &  969 &           889 &  969 &       901 &  947 &  \textbf{845} &  949 &  \textbf{845} &  845 \\
Edata12 &  900 &           912 &  912 &      1000 &  946 &  \textbf{878} &  915 &  \textbf{878} &  878 \\
Edata13 &  937 &           929 &  929 &       937 &  886 &  \textbf{866} &  937 &  \textbf{866} &  866 \\
Edata14 & 1221 &          1188 & 1157 &      1209 & 1213 & \textbf{1103} & 1176 & \textbf{1103} & 1103 \\
Edata15 & 1001 &           984 &  979 &      1017 & 1065 &  \textbf{960} &  987 &  \textbf{960} &  960 \\
Edata16 & 1142 &          1150 & 1061 &      1197 & 1164 & \textbf{1053} & 1150 & \textbf{1053} & 1053 \\
Edata17 & 1231 &          1240 & 1195 &      1195 & 1234 & \textbf{1123} & 1163 & \textbf{1123} & 1123 \\
Edata18 & 1356 &          1263 & 1397 &      1384 & 1297 & \textbf{1111} & 1312 & \textbf{1111} & 1111 \\
Edata19 & 1038 &          1012 & 1017 &      1107 & 1051 &  \textbf{892} &  986 &  \textbf{892} &  892 \\
Edata20 &  813 &           833 &  821 &       874 &  773 &  \textbf{707} &  765 &  \textbf{707} &  707 \\
Edata21 &  978 &           913 &  919 &       912 &  934 &  \textbf{842} &  915 &  \textbf{842} &  842 \\
Edata22 & 1011 &           964 &  941 &       950 &  972 &  \textbf{796} &  894 &  \textbf{796} &  796 \\
Edata23 &  989 &          1046 & 1007 &      1016 & 1004 &  \textbf{857} &  937 &  \textbf{857} &  857 \\
Edata24 & 1223 &          1200 & 1197 &      1199 & 1256 &          1083 & 1205 & \textbf{1063} & 1017 \\
Edata25 &  986 &          1049 & 1017 &      1020 & 1070 &  \textbf{924} & 1043 &           945 &  882 \\
Edata26 & 1074 &          1055 & 1122 &      1134 & 1137 & \textbf{1014} & 1116 &          1032 &  950 \\
Edata27 & 1056 &          1049 & 1025 &      1042 & 1086 &           962 & 1000 &  \textbf{955} &  909 \\
Edata28 & 1166 &          1070 & 1117 &      1111 & 1102 &  \textbf{937} & 1088 &           995 &  941 \\
Edata29 & 1272 &          1395 & 1394 &      1392 & 1348 &          1262 & 1230 & \textbf{1176} & 1125 \\
Edata30 & 1493 &          1390 & 1344 &      1557 & 1423 &          1363 & 1428 & \textbf{1249} & 1186 \\
Edata31 & 1456 &          1333 & 1331 &      1392 & 1344 & \textbf{1276} & 1368 &          1307 & 1149 \\
Edata32 & 1348 &          1378 & 1289 &      1317 & 1359 & \textbf{1208} & 1282 &          1244 & 1118 \\
Edata33 & 1470 &          1417 & 1371 &      1477 & 1410 & \textbf{1296} & 1358 &          1325 & 1204 \\
Edata34 & 1766 &          1693 & 1698 &      1753 & 1755 &          1806 & 1699 & \textbf{1653} & 1539 \\
Edata35 & 1903 &          1839 & 1872 &      1904 & 1904 &          1918 & 1832 & \textbf{1771} & 1698 \\
Edata36 & 1723 &          1710 & 1664 &      1812 & 1785 &          1892 & 1725 & \textbf{1629} & 1547 \\
Edata37 & 1772 &          1767 & 1778 &      1914 & 1817 &          1886 & 1719 & \textbf{1664} & 1604 \\
Edata38 & 1958 & \textbf{1858} & 1935 &      1983 & 2080 &          1971 & 1912 &          1897 & 1736 \\
Edata39 & 1304 &          1460 & 1347 &      1338 & 1295 & \textbf{1222} & 1368 &          1291 & 1162 \\
Edata40 & 1591 &          1557 & 1588 &      1578 & 1531 & \textbf{1425} & 1507 &          1476 & 1397 \\
Edata41 & 1382 &          1408 & 1444 &      1395 & 1370 & \textbf{1179} & 1369 &          1278 & 1144 \\
Edata42 & 1414 &          1400 & 1414 &      1406 & 1412 & \textbf{1215} & 1395 &          1288 & 1184 \\
Edata43 & 1321 &          1306 & 1316 &      1446 & 1311 & \textbf{1177} & 1338 &          1239 & 1150 \\
\hline
Gap (\%) & $15.53 $ & $ 13.73 $ & $ 13.2 $ & $ 179 $ & $ 15.54 $ & $ 4.4 $ & $ 122  $ & $ \bm{ 3.78 }$ \\
\end{tabular}
\end{table}
\end{tiny}

\begin{tiny}\setlength\tabcolsep{4pt}
\begin{table}[H]
\centering
\small
\caption{Comparison of the best makespan obtained by the methods proposed by \cite{song2022flexible},  \cite{wang2023flexible}, \cite{ho2024residual}, \cite{yuan2023solving}, \cite{echeverria2023solving}, OR-Tools, BC and BCxCP in the rdata dataset.} \label{tab:rdataall} 
\begin{tabular}{l|r|r|r|r|r|r|r|r|r}
Name &           Song &           Wang &             Ho &     Echeverria &  Yuan &            OR &             BC &           BCxCP &  UB \\
\hline
\hline
Rdata04 &  603 &  617 &  618 &       617 &  663 &          587 &  614 &  \textbf{578} &  571 \\
Rdata05 &  611 &  615 &  598 &       609 &  666 &          538 &  587 &  \textbf{532} &  530 \\
Rdata06 &  520 &  513 &  516 &       561 &  525 & \textbf{479} &  508 &  \textbf{479} &  478 \\
Rdata07 &  586 &  551 &  546 &       540 &  573 &          509 &  574 &  \textbf{508} &  502 \\
Rdata08 &  489 &  511 &  502 &       489 &  500 & \textbf{462} &  481 &  \textbf{462} &  457 \\
Rdata09 &  815 &  822 &  810 &       882 &  856 &          803 &  830 &  \textbf{801} &  799 \\
Rdata10 &  774 &  790 &  782 &       795 &  825 & \textbf{751} &  783 &           753 &  750 \\
Rdata11 &  837 &  829 &  789 &       859 &  820 & \textbf{768} &  795 &           770 &  765 \\
Rdata12 &  898 &  883 &  882 &       894 &  885 &          857 &  878 &  \textbf{856} &  853 \\
Rdata13 &  828 &  870 &  840 &       838 &  870 &          816 &  853 &  \textbf{807} &  804 \\
Rdata14 & 1089 & 1133 & 1094 &      1094 & 1131 &         1086 & 1091 & \textbf{1071} & 1071 \\
Rdata15 &  949 &  965 &  987 &       974 &  977 &          939 &  957 &  \textbf{936} &  936 \\
Rdata16 & 1091 & 1135 & 1067 &      1056 & 1051 &         1055 & 1050 & \textbf{1039} & 1038 \\
Rdata17 & 1138 & 1176 & 1085 &      1127 & 1110 &         1085 & 1093 & \textbf{1072} & 1070 \\
Rdata18 & 1197 & 1224 & 1151 &      1178 & 1236 &         1124 & 1122 & \textbf{1091} & 1090 \\
Rdata19 &  825 &  829 &  784 &       867 &  821 & \textbf{717} &  776 &  \textbf{717} &  717 \\
Rdata20 &  758 &  846 &  732 &       767 &  720 & \textbf{646} &  671 &  \textbf{646} &  646 \\
Rdata21 &  730 &  759 &  772 &       771 &  766 & \textbf{666} &  784 &           703 &  666 \\
Rdata22 &  973 &  874 &  828 &       923 &  812 &          719 &  821 &  \textbf{717} &  700 \\
Rdata23 &  891 &  858 &  889 &       941 &  878 & \textbf{756} &  880 &           777 &  756 \\
Rdata24 &  993 &  991 &  987 &       978 &  967 &          960 &  948 &  \textbf{865} &  835 \\
Rdata25 &  918 &  923 &  894 &       892 &  893 &          983 &  889 &  \textbf{814} &  760 \\
Rdata26 &  921 &  947 &  983 &      1023 & 1010 &          965 &  966 &  \textbf{909} &  842 \\
Rdata27 &  948 &  947 &  896 &       977 &  928 &          957 &  908 &  \textbf{870} &  808 \\
Rdata28 &  943 &  977 &  935 &       862 &  939 &          953 &  916 &  \textbf{839} &  791 \\
Rdata29 & 1173 & 1154 & 1169 &      1177 & 1175 &         1312 & 1122 & \textbf{1100} & 1061 \\
Rdata30 & 1207 & 1211 & 1209 &      1295 & 1214 &         1295 & 1230 & \textbf{1136} & 1091 \\
Rdata31 & 1202 & 1214 & 1196 &      1218 & 1175 &         1328 & 1154 & \textbf{1109} & 1080 \\
Rdata32 & 1082 & 1082 & 1106 &      1115 & 1111 &         1288 & 1069 & \textbf{1027} &  998 \\
Rdata33 & 1193 & 1280 & 1191 &      1208 & 1275 &         1383 & 1186 & \textbf{1128} & 1078 \\
Rdata34 & 1593 & 1573 & 1576 &      1632 & 1636 &         1910 & 1563 & \textbf{1532} & 1521 \\
Rdata35 & 1748 & 1720 & 1742 &      1741 & 1765 &         2041 & 1741 & \textbf{1683} & 1659 \\
Rdata36 & 1558 & 1547 & 1529 &      1591 & 1542 &         1804 & 1542 & \textbf{1512} & 1499 \\
Rdata37 & 1570 & 1579 & 1601 &      1622 & 1686 &         1797 & 1565 & \textbf{1540} & 1536 \\
Rdata38 & 1636 & 1666 & 1653 &      1628 & 1706 &         1864 & 1651 & \textbf{1560} & 1550 \\
Rdata39 & 1204 & 1162 & 1192 &      1226 & 1229 &         1327 & 1154 & \textbf{1127} & 1030 \\
Rdata40 & 1219 & 1229 & 1208 &      1299 & 1294 &         1205 & 1184 & \textbf{1135} & 1077 \\
Rdata41 & 1127 & 1230 & 1104 &      1202 & 1107 &         1159 & 1080 & \textbf{1047} &  962 \\
Rdata42 & 1240 & 1207 & 1222 &      1205 & 1228 &         1262 & 1132 & \textbf{1091} & 1024 \\
Rdata43 & 1156 & 1168 & 1082 &      1183 & 1108 &         1089 & 1080 & \textbf{1050} &  970 \\
\hline
Gap (\%) & $ 11.15  $ & $ 12.17   $ & $ 9.65   $ & $ 12.78   $ & $ 12.14  $ & $ 11.1  $ & $ 8.26   $ & $ \bm{ 2.75 }$ \\
\end{tabular}
\end{table}
\end{tiny}

\begin{table}[H]
\centering
\caption{Comparison of the best makespan obtained by the methods proposed by \cite{song2022flexible},  \cite{wang2023flexible}, \cite{ho2024residual}, \cite{yuan2023solving}, \cite{echeverria2023solving}, OR-Tools, BC and BCxCP in the Brandimarte dataset.} \label{tab:brandimarteall} 
\begin{tabular}{l|r|r|r|r|r|r|r|r|r}
Name &           Song &           Wang &             Ho &     Echeverria &  Yuan &            OR &             BC &           BCxCP &  UB \\
\hline
\hline
Mk01 &           43 &           43 &           45 &           44 &  44 & \textbf{40} &           41 &  \textbf{40} &  39 \\
Mk02 &           38 &           29 &           29 &           40 &  31 & \textbf{28} &           31 &           29 &  26 \\
Mk03 & \textbf{204} & \textbf{204} & \textbf{204} & \textbf{204} & 211 &         229 & \textbf{204} & \textbf{204} & 204 \\
Mk04 &           81 &           69 &           67 &           75 &  78 & \textbf{62} &           72 &           64 &  60 \\
Mk05 &          192 & \textbf{176} &          185 &          189 & 183 &         220 &          178 & \textbf{176} & 172 \\
Mk06 &           99 &           86 &           74 &          104 &  74 &          96 &           77 &  \textbf{73} &  58 \\
Mk07 &          221 & \textbf{148} &          150 &          211 & 156 &         178 &          167 &          156 & 139 \\
Mk08 &          529 & \textbf{523} & \textbf{523} & \textbf{523} & 524 &         590 &          527 & \textbf{523} & 523 \\
Mk09 &          339 &          335 &          314 &          337 & 326 &         503 &          311 & \textbf{307} & 307 \\
Mk10 &          264 & 230 &          \textbf{225} &          258 & 241 &         383 &          245 &          236 & 197 \\
\hline
Gap (\%) & $27.83  $ & $ 11.97 $ & $ 9.81 $ & $ 27.34 $ & $ 14.04  $ & $ 31.84 $ & $ 12.72  $ & $ \bm{ 8.1}$ &
\end{tabular}
\end{table}

\begin{tiny}\setlength\tabcolsep{4pt}
\begin{table}[H]
\centering
\caption{Comparison of the best makespan obtained by the methods proposed by \cite{song2022flexible},  \cite{wang2023flexible}, \cite{yuan2023solving}, \cite{echeverria2023solving}, OR-Tools, BC and BCxCP in the Dauzere dataset.} \label{tab:dpaullieall} 
\begin{tabular}{l|r|r|r|r|r|r|r|r}
Name &           Song &           Wang &                  Echeverria &  Yuan &            OR &             BC &           BCxCP &  UB \\
\hline
\hline
01a &          2836 & 2865 &           2911 & 2947 & \textbf{2686} & 2922 &          2738 & 2518 \\
02a &          2579 & 2533 &           2472 & 2495 &          2685 & 2351 & \textbf{2308} & 2231 \\
03a &          2283 & 2306 &           2269 & 2320 &          2938 & 2280 & \textbf{2242} & 2229 \\
04a &          2913 & 2879 &           2918 & 3037 &          2769 & 2718 & \textbf{2656} & 2503 \\
05a &          2470 & 2415 &           2403 & 2496 &          2824 & 2342 & \textbf{2296} & 2216 \\
06a &          2260 & 2331 &           2253 & 2286 &          2546 & 2240 & \textbf{2221} & 2196 \\
07a &          2829 & 2643 &           2804 & 2820 &          2910 & 2741 & \textbf{2608} & 2283 \\
08a &          2190 & 2230 &           2138 & 2259 &          3156 & 2161 & \textbf{2110} & 2069 \\
09a & \textbf{2073} & 2104 &           2144 & 2155 &          2764 & 2156 &          2155 & 2066 \\
10a &          2776 & 2664 &           2686 & 2721 &          2910 & 2641 & \textbf{2588} & 2291 \\
11a &          2172 & 2210 &           2228 & 2206 &          2751 & 2157 & \textbf{2130} & 2063 \\
12a &          2103 & 2094 &           2088 & 2157 &          2696 & 2124 & \textbf{2082} & 2030 \\
13a &          2674 & 2622 &           2681 & 2755 &          3008 & 2623 & \textbf{2535} & 2257 \\
14a &          2210 & 2224 &           2316 & 2277 &          2984 & 2261 & \textbf{2193} & 2167 \\
15a &          2200 & 2197 &  \textbf{2186} & 2202 &          3366 & 2202 &          2192 & 2165 \\
16a &          2631 & 2737 &           2683 & 2680 &          2996 & 2573 & \textbf{2511} & 2255 \\
17a &          2214 & 2207 &           2272 & 2309 &          2983 & 2206 & \textbf{2186} & 2140 \\
18a &          2190 & 2172 &           2178 & 2216 &          3186 & 2174 & \textbf{2160} & 2127 \\
\hline
Gap (\%) & $ 9.26 $ & $ 8.88  $  & $ 9.33   $ & $ 11.1 $ & $ 31.57  $ & $ 7.5 $ & $ \bm{ 5.15  }$ 
\end{tabular}
\end{table}
\end{tiny}

\end{document}